\documentclass{article}
\usepackage{ijcai19}

\usepackage{booktabs}

\usepackage{times}
\usepackage{soul}
\usepackage{url}
\usepackage[hidelinks]{hyperref}
\usepackage[utf8]{inputenc}
\usepackage[small]{caption}
\usepackage{graphicx}
\usepackage{booktabs}
\usepackage{algorithm}
\usepackage[noend]{algorithmic}
\usepackage{hyperref}
\usepackage{url}

\usepackage[dvipsnames]{xcolor}

\usepackage{nicefrac}       
\usepackage{microtype}      

\usepackage{subfigure}
\usepackage{amsmath, amsthm, amssymb}

\newcommand{\figwidththree}{0.31\textwidth}
\newcommand{\figwidthfour}{0.23\textwidth}
\newcommand{\figwidthfive}{0.18\textwidth}
\newcommand{\figwidthsix}{0.15\textwidth}

\newcommand{\Actions}{\mathcal{A}}
\newcommand{\States}{\mathcal{S}}

\newcommand{\defeq}{\mathrel{\overset{\makebox[0pt]{\mbox{\normalfont\tiny\sffamily def}}}{=}}}

\newcommand{\Sigmamat}{\boldsymbol{\Sigma}}

\newcommand{\gvec}{\mathbf{g}}

\newcommand{\svec}{\mathbf{s}}

\newcommand{\bsc}{B_{sc}}
\newcommand{\ber}{B_{er}}

\newcommand{\eye}{\mathbf{I}}

\newcommand{\RR}{\mathbb{R}}

\newcommand{\covmathat}{\hat{\Sigmamat}_s}
\DeclareMathOperator*{\argmax}{arg\,max}

\newif\ifconsiderlater
\considerlaterfalse

\ifconsiderlater
	\newcommand{\todo}[1]{{\textbf{XXX [#1] XXX}}}
\else
\newcommand{\todo}[1]{}
\fi

\newcommand{\todoNew}[1]{} 

\newif\ifSupp
\Supptrue

\title{Hill Climbing on Value Estimates for Search-control in Dyna}

\author{
	Yangchen Pan$^1$\and
	Hengshuai Yao$^2$\and
	Amir-massoud Farahmand$^{3,4}$\And
	Martha White$^1$
	\affiliations
	$^1$Department of Computing Science, University of Alberta, Canada\\
	$^2$Huawei HiSilicon, Canada\\
	$^3$Vector Institute, Canada\\
	$^4$Department of Computer Science, University of Toronto, Canada
	\emails
	pan6@ualberta.ca,
	hengshuai.yao@huawei.com,
	farahmand@vectorinstitute.ai,
	whitem@ualberta.ca
}

\begin{document}

\maketitle
\begin{abstract}
Dyna is an architecture for model-based reinforcement learning (RL), where simulated experience from a model is used to update policies or value functions. A key component of Dyna is search-control, the mechanism to generate the state and action from which the agent queries the model, which remains largely unexplored. In this work, we propose to generate such states by using the trajectory obtained from Hill Climbing (HC) the current estimate of the value function. This has the effect of propagating value from high-value regions and of preemptively updating value estimates of the regions that the agent is likely to visit next. We derive a noisy projected natural gradient algorithm for hill climbing, and highlight a connection to Langevin dynamics. We provide an empirical demonstration on four classical domains that our algorithm, HC-Dyna, can obtain significant sample efficiency improvements. We study the properties of different sampling distributions for search-control, and find that there appears to be a benefit specifically from using the samples generated by climbing on current value estimates from low-value to high-value region.
\end{abstract}

\section{Introduction}
\label{sec:HC-Dyna-Introduction}

\todoNew{Do we refer to the appendix/supplementary material at any point?}

\todoNew{I believe IJCAI allows buying extra pages. Do we want to include the appendix?}

Experience replay (ER) \cite{lin1992self} is currently the most common way to train value functions approximated as neural networks (NNs), in an online RL setting \cite{AdamBusoniuBabuska2012,wawrzynski2013autonomous}. The buffer in ER is typically a recency buffer, storing the most recent transitions, composed of state, action, next state and reward. At each environment time step, the NN gets updated by using a mini-batch of samples from the ER buffer, that is, the agent replays those transitions. 
ER enables the agent to be more sample efficient, and in fact can be seen as a simple form of model-based RL \cite{vanseijen2015adeeper}. This connection is specific to the Dyna architecture \cite{sutton1990integrated,sutton1991dyna}, where the agent maintains a search-control (SC) queue of pairs of states and actions and uses a model to generate next states and rewards. These simulated transitions are used to update values. ER, then, can be seen as a variant of Dyna with a nonparameteric model, where search-control is determined by the observed states and actions. 

By moving beyond ER to Dyna with a learned model, we can potentially benefit from increased flexibility in obtaining simulated transitions. 
Having access to a model allows us to generate unobserved transitions, from a given state-action pair. For example, a model allows the agent to obtain on-policy or exploratory samples from a given state, which has been reported to have advantages~\cite{gu2016continuous,yangchen2018rem,SantosMartin2012,Peng2018DeepDynaQ}. More generally, models allow for a variety of choices for search-control, which is critical as it emphasizes different states during the planning phase.
Prioritized sweeping \cite{moore1993prioritized} uses the model to obtain predecessor states, with states sampled according to the absolute value of temporal difference error. This early work, and more recent work \cite{sutton2008dyna,yangchen2018rem,dane2018mbrltabulation}, showed this addition significantly outperformed Dyna with states uniformly sampled from observed states. 
Most of the work on search-control, however, has been limited to sampling visited or predecessor states. Predecessor states require a reverse model, which can be limiting. The range of possibilities has yet to be explored for search-control and there is room for many more ideas. 

In this work, we investigate using sampled trajectories by hill climbing on our learned value function to generate states for search-control. Updating along such trajectories has the effect of propagating value from regions the agent currently believes to be high-value. This strategy enables the agent to preemptively update regions where it is likely to visit next. Further, it focuses updates in areas where approximate values are high, and so important to the agent.
To obtain such states for search-control, we propose a noisy natural projected gradient algorithm. 
We show this has a connection to Langevin dynamics, whose distribution converges to the Gibbs distribution,  where the density is proportional to the exponential of the state values. We empirically study different sampling distributions for populating the search-control queue, and verify the effectiveness of hill climbing based on estimated values. We conduct experiments showing improved performance in four benchmark domains, as compared to DQN\footnote{We use DQN to refer to the algorithm by~\cite{mnih2015humanlevel} that uses ER and target network, but not the exact original architecture.}, and illustrate the usage of our architecture for continuous control. 


\section{Background}
\label{sec:HC-Dyna-RL-Background}

We formalize the environment as a Markov Decision Process (MDP) 
 $(\States, \Actions, \mathbb{P}, R,  \gamma)$, where
$\States$ is the state space, $\Actions$ is the action space, $\mathbb{P} : \States \times \Actions \times \States \rightarrow [0, 1]$ is the transition probabilities, $R:  \States \times \Actions \times \States \rightarrow \RR$ is the reward function, and $\gamma \in [0,1]$ is the discount factor.
At each time step $t = 1, 2, \dotsc$, the agent observes a state $s_t \in \States$ and takes an action $a_t \in \Actions$, transitions to $s_{t+1} \sim \mathbb{P}(\cdot| s_t, a_t)$ and receives a scalar reward $r_{t+1} \in \RR$ according to the reward function $R$.

Typically, the goal is to learn a policy to maximize the expected return starting from some fixed initial state. One popular algorithm is Q-learning, by which we can obtain approximate action-values $Q_\theta: \States \times \Actions \rightarrow \RR$ for parameters $\theta$. The policy corresponds to acting greedily according to these action-values: for each state, select action $\arg\max_{a \in \Actions} Q(s,a)$. The Q-learning update for a sampled transition $s_t, a_t, r_{t+1}, s_{t+1}$ is  
\begin{align*}
\setlength{\abovedisplayskip}{4pt}
\theta &\gets \theta + \alpha \delta_t \nabla Q_\theta(s_t, a_t)\\
&\text{where } \delta_t \defeq r_{t+1} + \max_{a' \in \Actions} Q_\theta(s_{t+1},a') - Q_\theta(s_t,a_t) 
\setlength{\belowdisplayskip}{4pt}
\end{align*} 
Though frequently used, such an update may not be sound with function approximation. Target networks \cite{mnih2015humanlevel} are typically used to improve stability when training NNs, where the bootstrap target on the next step is a fixed, older estimate of the action-values. \todo{We should mention that this is Fitted Value/Q Iteration algorithm. Szepesvari \& Munos, Ernst 2005 -AMF}

ER and Dyna can both be used to improve sample efficiency of DQN. Dyna is a model-based method that simulates (or replays) transitions, to reduce the number of required interactions with the environment. A model is sometimes available a priori (e.g., from physical equations of the dynamics) or is learned using data collected through interacting with the environment. The generic Dyna architecture, with explicit pseudo-code given by \cite[Chapter 8]{SuttonBarto2018}, can be summarized as follows.
When the agent interacts with the real world, it updates both the action-value function and the learned model using the real transition. 
The agent then performs $n$ planning steps. In each planning step, the agent samples $(\tilde{s}, \tilde{a})$ from the search-control queue, generates next state $\tilde{s}'$ and reward $\tilde{r}$ from $(\tilde{s}, \tilde{a})$ using the model, and updates the action-values using Q-learning with the tuple $(\tilde{s}, \tilde{a}, \tilde{r}, \tilde{s}')$.


\section{A Motivating Example}
\label{sec:HC-Dyna-RL-ClimbingValueFunctionSurface}

In this section we provide an example of how the value function surface changes during learning on a simple continuous-state GridWorld domain.
This provides intuition on why it is useful to populate the search-control queue with states obtained by hill climbing on the estimated value function, as proposed in the next section.

\begin{figure*}[t]
	\centering
	\subfigure[GridWorld domain]{
			\includegraphics[width=\figwidthfive]{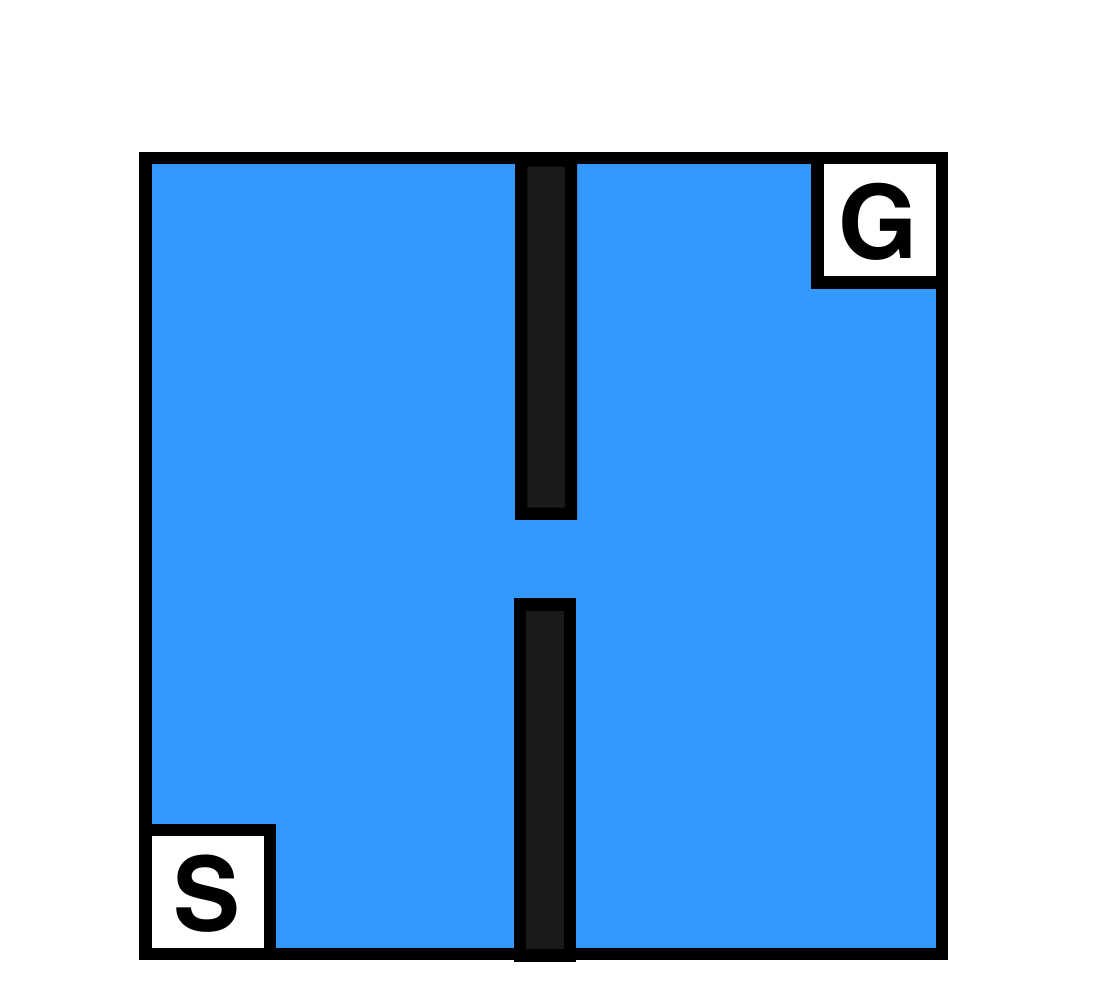} \label{fig:gridworld}}
	\subfigure[Before update]{
			\includegraphics[width=\figwidthfive]{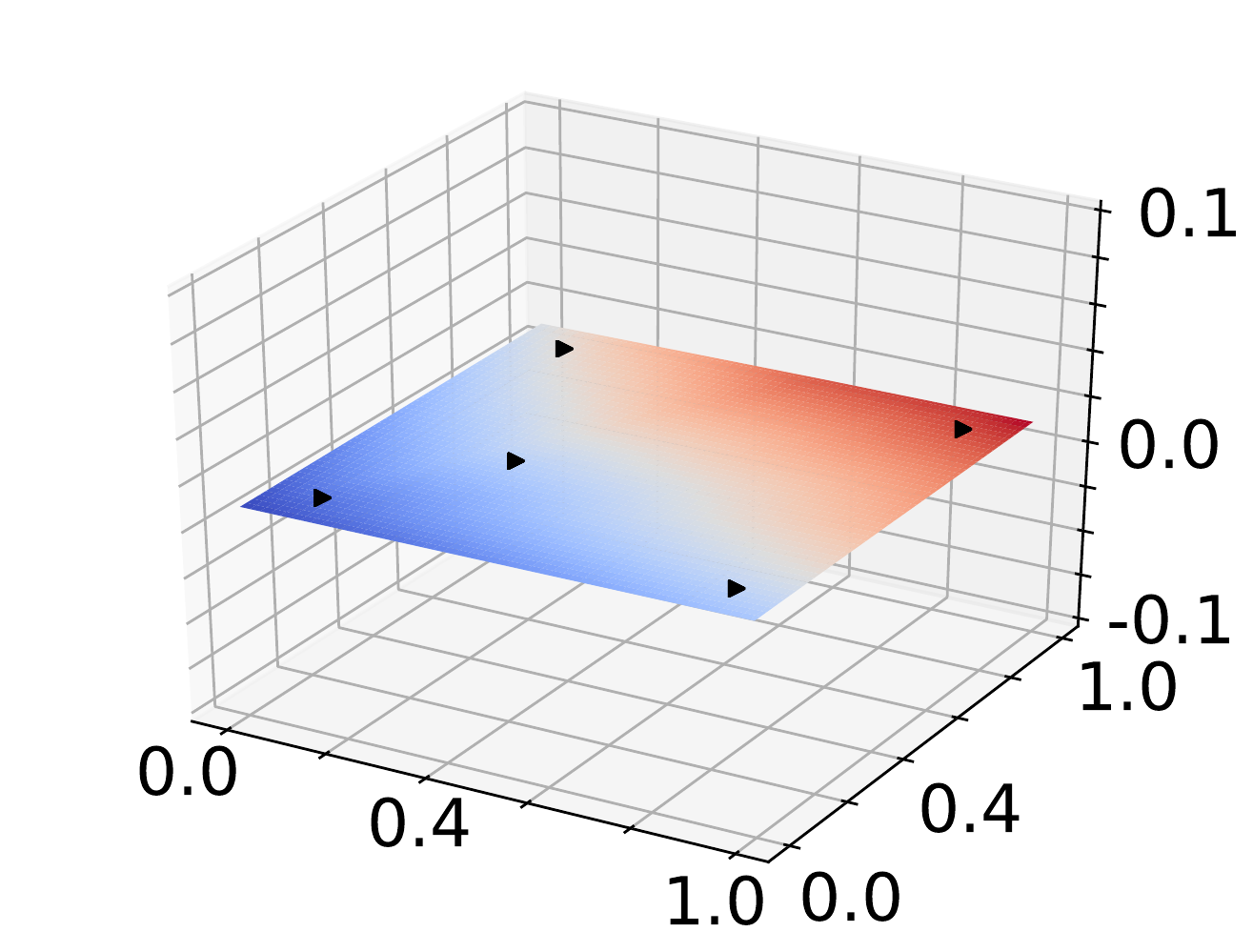} \label{fig:travel0}}
	\subfigure[Update 14k times]{
			\includegraphics[width=\figwidthfive]{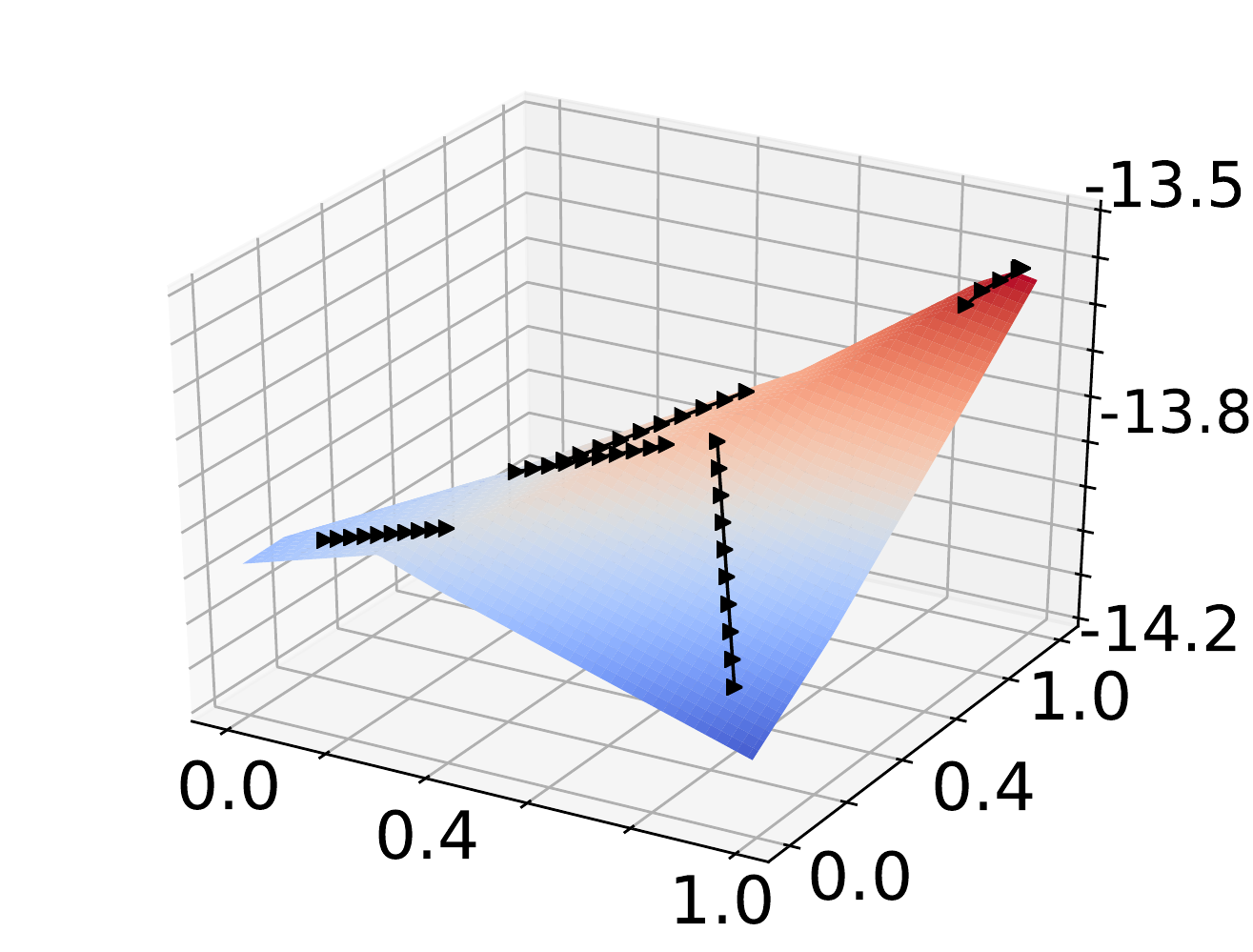} \label{fig:travel14k}}
	\subfigure[Update 20k times]{
			\includegraphics[width=\figwidthfive]{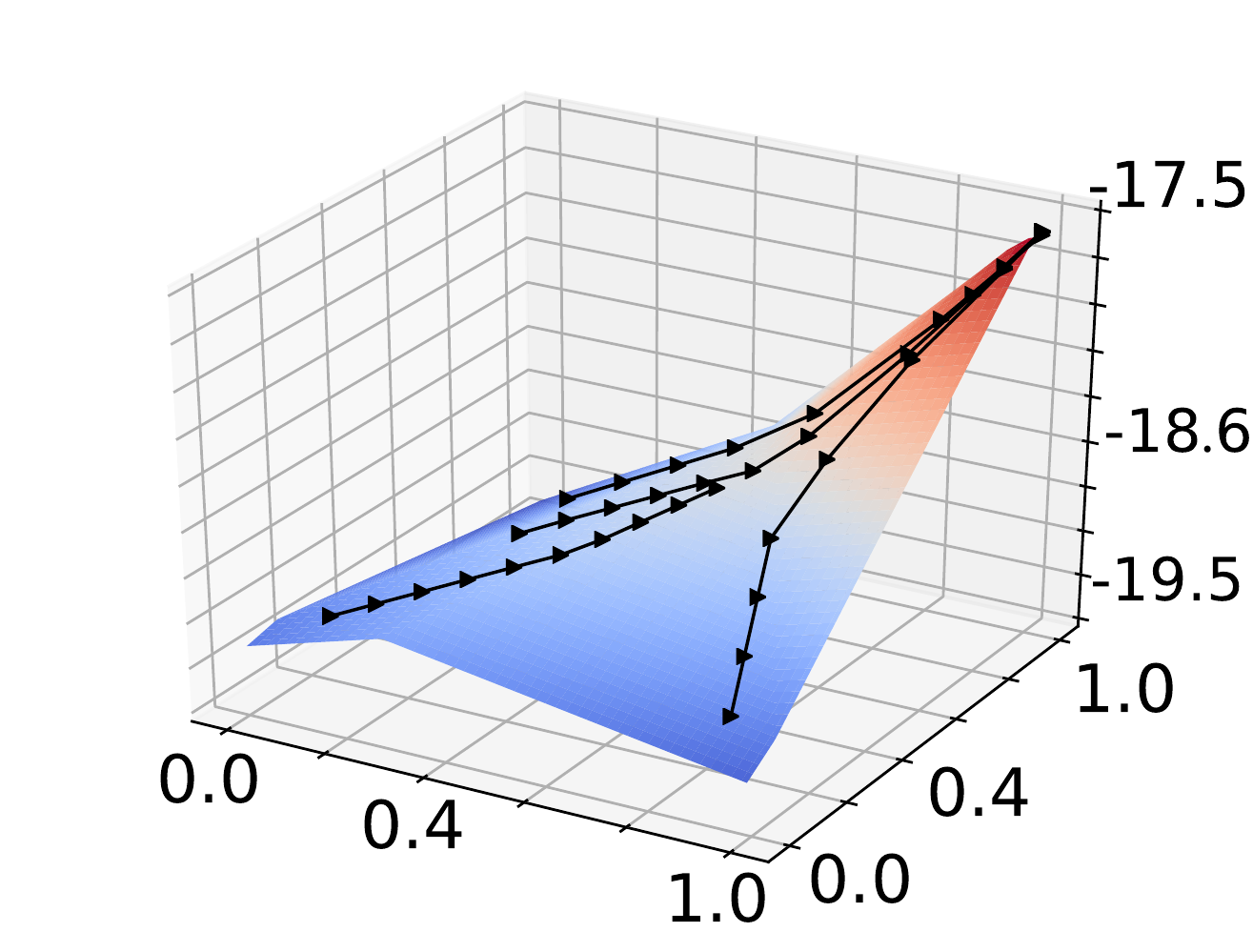} \label{fig:travel20k}}
	\subfigure[Negation of NN]{
				\includegraphics[width=\figwidthfive]{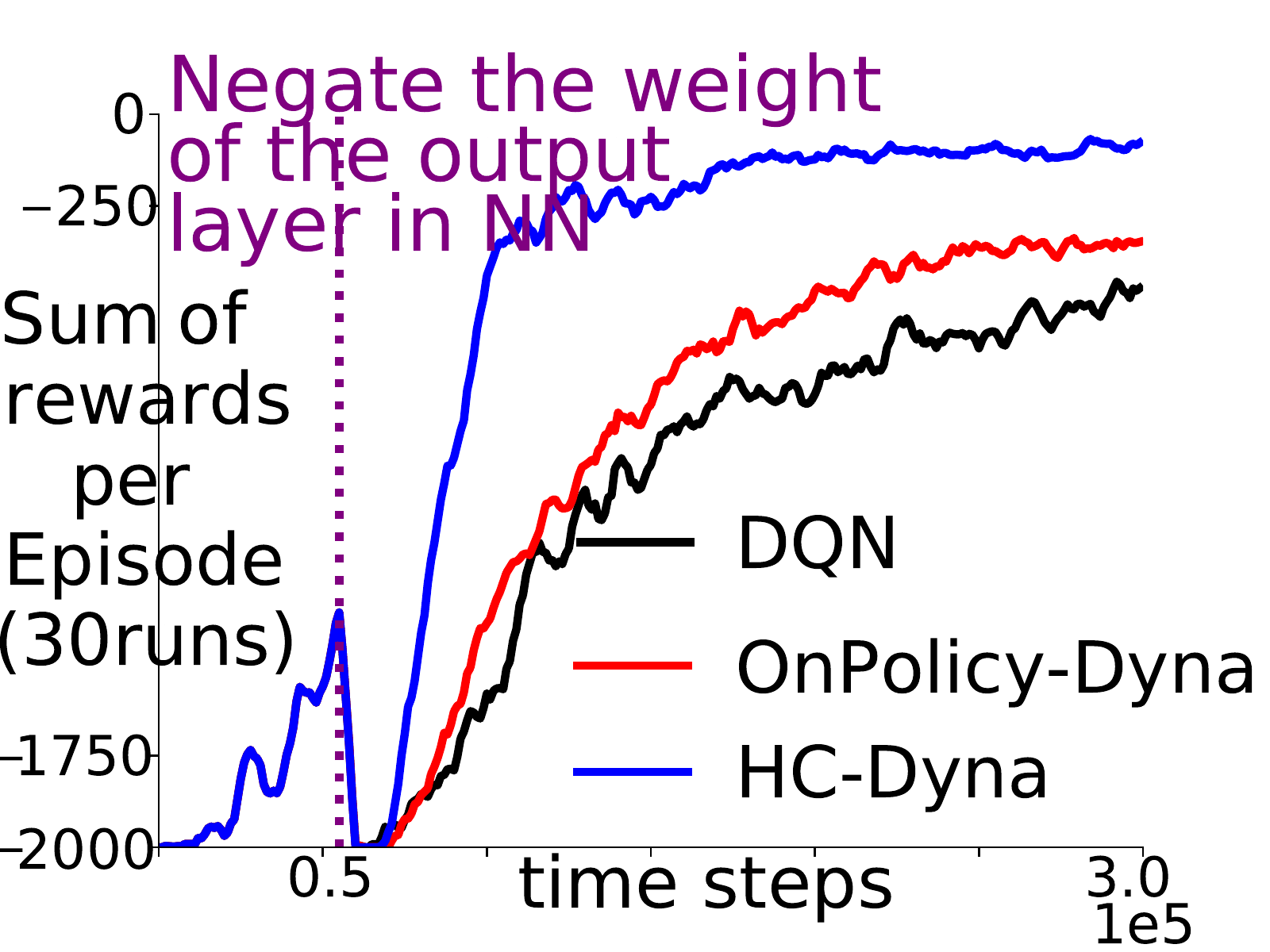} \label{fig:reverse}}
	\caption{
		(b-d) The value function on the GridWorld domain with gradient ascent trajectories. (e) shows learning curves (sum of rewards per episode v.s. time steps) where each algorithm needs to recover from a bad NN initialization (i.e. the value function looks like the reverse of (d)). 
		}\label{fig:gridworldexample}
	\vspace{-0.25cm}
\end{figure*}

Consider the GridWorld in Figure~\ref{fig:gridworld}, which is a variant of the one introduced by \cite{peng1993efficient}. In each episode, the agent starts from a uniformly sampled point from the area $[0,0.05]^2$ and  terminates when it reaches the goal area $[0.95, 1.0]^2$. There are four actions $\{\textsc{up}, \textsc{down}, \textsc{left}, \textsc{right}\}$; each leads to a $0.05$ unit move towards the corresponding direction. As a cost-to-goal problem, the reward is $-1$ per step.

In Figure~\ref{fig:gridworldexample}, we plot the value function surface after $0$, $14$k and $20$k mini-batch updates to DQN. We visualize the gradient ascent trajectories with $100$ gradient steps starting from five states $(0.1,0.1)$, $(0.9,0.9)$, $(0.1,0.9)$, $(0.9,0.1)$, and $(0.3,0.4)$. The gradient of the value function used in the gradient ascent is
\begin{equation}
\setlength{\abovedisplayskip}{3pt}
\label{eq:grad-of-V}
\nabla_s V(s) = 
\nabla_s \max_a Q_\theta (s, a),
\setlength{\belowdisplayskip}{4pt}
\end{equation}
At the beginning, with a randomly initialized NN, the gradient with respect to state is almost zero, as seen in Figure~\ref{fig:travel0}.
As the DQN agent updates its parameters, the gradient ascent generates trajectories directed towards the goal, though after only 14k steps, these are not yet contiguous, as seen Figure~\ref{fig:travel14k}. 
After $20$k steps, as in Figure~\ref{fig:travel20k}, even though the value function is still inaccurate, the gradient ascent trajectories take all initial states to the goal area.
This suggests that as long as the estimated value function roughly reflects the shape of the optimal value function, the trajectories provide a demonstration of how to reach the goal---or high-value regions---and speed up learning by focusing updates on these relevant regions. 

More generally, by focusing planning on regions the agent \emph{thinks} are high-value, it can quickly correct value function estimates before visiting those regions, and so avoid unnecessary interaction. We demonstrate this in Figure~\ref{fig:reverse}, where the agent obtains gains in performance by updating from high-value states, even when its value estimates have the wrong shape. After 20k learning steps, the values are flipped by negating the sign of the parameters in the output layer of the NN. HC-Dyna, introduced in Section \ref{sec:HC-Dyna-RL-Algorithm}, quickly recovers compared to DQN and OnPolicy updates from the ER buffer. 
Planning steps help pushing down these erroneously high-values, and the agent can recover much more quickly. \todo{Don't we have too many quickly here? -AMF}

\section{Effective Hill Climbing}

To generate states for search control, we need an algorithm that can climb on the estimated value function surface. For general value function approximators, such as NNs, this can be difficult. The value function surface can be very flat or very rugged, causing the gradient ascent to get stuck in local optima and hence interrupt the gradient traveling process. Further, the state variables may have very different numerical scales. When using a regular gradient ascent method, it is likely for the state variables with a smaller numerical scale to immediately go out of the state space. Lastly, gradient ascent is unconstrained, potentially generating unrealizable states. 

In this section, we propose solutions for all these issues. We provide a noisy invariant projected gradient ascent strategy to generate meaningful trajectories of states for search-control. We then discuss connections to Langevin dynamics, a model for heat diffusion, which provides insight into the sampling distribution of our search-control queue. 
\todo{Isn't it the case that we only use the projected version in some of the experiments? -AMF}

\subsection{Noisy Natural Projected Gradient Ascent}

To address the first issue, of flat or rugged function surfaces, we propose to add Gaussian noise on each gradient ascent step. Intuitively, this provides robustness to flat regions and avoids getting stuck in local maxima on the function surface, by diffusing across the surface to high-value regions. \todo{On a flat surface, adding a Gaussian noise leads to a Brownian motion. I think a Brownian motion is not the most efficient way of getting out of a flat region. -AMF}
\todo{Have we decided to remove the connection to Langevin? -AMF}
 
To address the second issue of vastly different numerical scales among state variables, we use a standard strategy to be invariant to scale: natural gradient ascent. 
A popular choice of natural gradient is derived by defining the metric tensor as the Fisher information matrix~\cite{amari1998whyng,amari1998ngworks,philip2016energeticng}. 
We introduce a simple and computationally efficient metric tensor: the inverse of covariance matrix of the states $\Sigmamat_\svec^{-1}$. This choice is simple, because the covariance matrix can easily be estimated online. 
We can define the following inner product:
\begin{equation*}\label{gametric}
\setlength{\abovedisplayskip}{3pt}
	\langle s, s' \rangle = s^\top \Sigmamat_s^{-1} s', \forall s, s' \in \States,
	\setlength{\belowdisplayskip}{3pt}
\end{equation*}
which induces a vector space---the Riemannian manifold---where we can compute the distance of two points $s$ and $s + \Delta$ that are close to each other by $d(s, s + \Delta) \defeq \Delta^\top \Sigmamat_s^{-1}  \Delta$. The steepest ascent updating rule based on this distance metric becomes 
$s \gets s + \alpha \Sigmamat_\svec \gvec$,
where $\gvec$ is the gradient vector. 

\begin{figure}[t!]
	\includegraphics[width=0.28\textwidth]{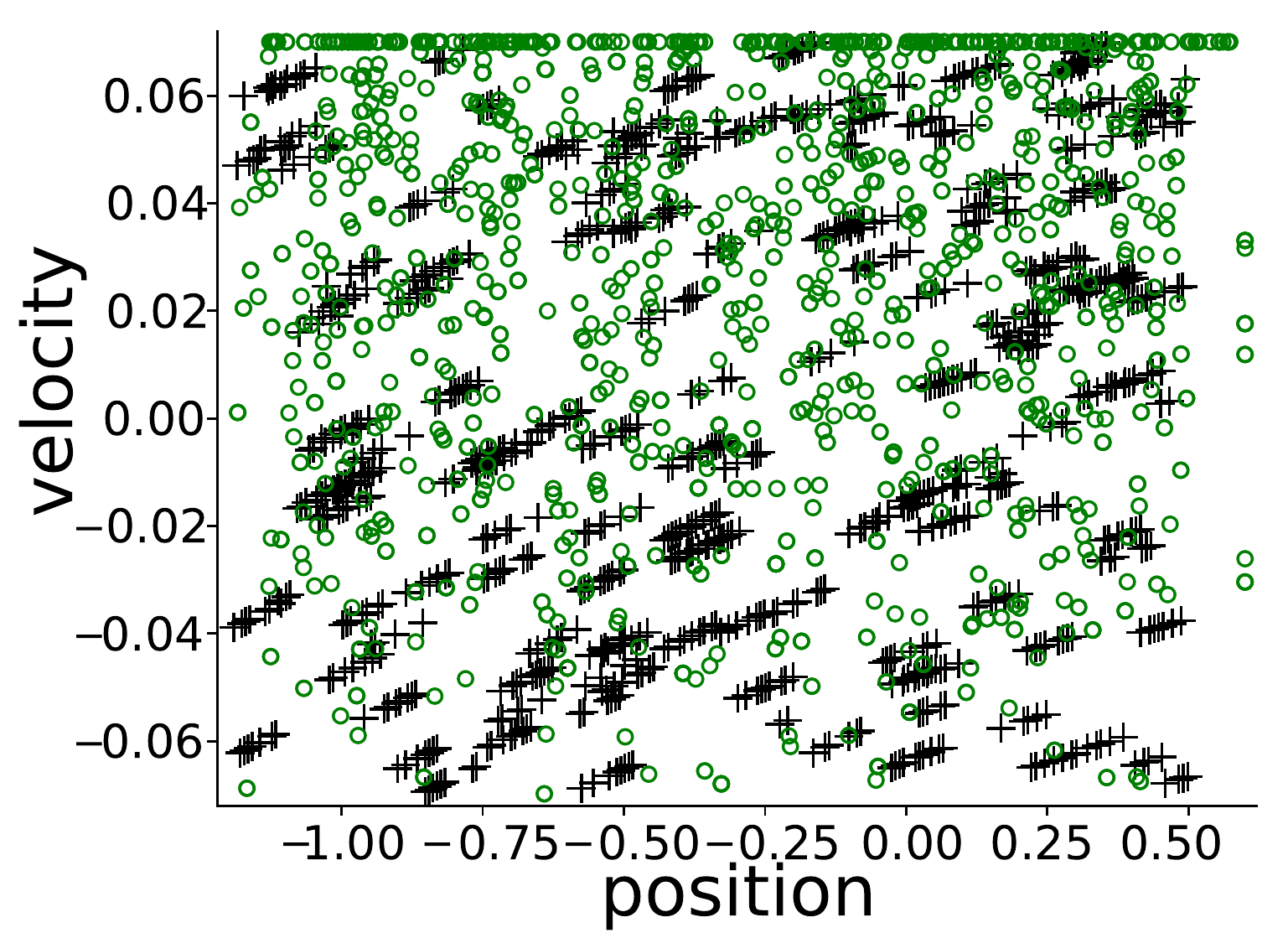}
		\caption{
			The search-control queue filled by using \textbf{({\color{black} +})} or not using \textbf{({\color{OliveGreen} o})} natural gradient on MountainCar-v0. 
		} \label{fig:ngd}
	\vspace{-0.25cm}	
\end{figure}

We demonstrate the utility of using the natural gradient scaling. Figure~\ref{fig:ngd} shows the states from the search-control queue filled by hill climbing in early stages of learning (after 8000 steps) on MountainCar. The domain has two state variables with very different numerical scale: position $\in [-1.2, 0.6]$ and velocity $\in [-0.07, 0.07]$. Using a regular gradient update, the queue shows a state distribution with many states concentrated near the top since it is very easy for the velocity variable to go out of boundary. In contrast, the one with natural gradient, shows clear trajectories with an obvious tendency to the right top area (position $\ge 0.5$), which is the goal area. 
\todo{Do you project the state variables to a certain range? -AMF No I do not. Projection to certain range is not general since some state variable is not bounded. 
	Q (AMF): So why do we observe a lot of circles concentrated very neatly at the top? Something should prevent them to move upwards. YC: that is the case when we do not use natural gradient, which is to contrast with our method. When not using natural gradient, it is very easy for the small scale variable to go out of boundary. }

We use projected gradient updates to address the third issue regarding unrealizable states.
We explain the issue and solution using the Acrobot domain. The first two state variables are $\cos \theta, \sin \theta$, where $\theta$ is the angle between the first robot arm's link and the vector pointing downwards. This induces the restriction that $\cos^2 \theta + \sin^2 \theta = 1$. The hill climbing process could generate many states that do not satisfy this restriction. This could potentially degrade performance, since the NN needs to generalize to these states unnecessarily. 
We can use a projection operator $\Pi$ to enforce such restrictions, whenever known, after each gradient ascent step. In Acrobot, $\Pi$ is a simple normalization. In many settings, the constraints are simple box constraints, with projection just inside the boundary. 

Now we are ready to introduce our final hill climbing rule:
\begin{equation}\label{gaupdaterule}
\setlength{\abovedisplayskip}{3pt}
s \gets  \Pi \left(s + \alpha \Sigmamat_s \gvec + \mathcal{N}\right),
\setlength{\belowdisplayskip}{3pt}
\end{equation}
where 
$\mathcal{N}$ is Gaussian noise and $\alpha$ a stepsize. For simplicity, we set the stepsize to $\alpha = 0.1/||\Sigmamat_\svec \gvec||$ across all results in this work, though of course there could be better choices. 
\subsection{Connection to Langevin Dynamics}\label{sec:HC-Dyna-theory} 
\todo{This section should be revised a bit to make it more aligned with our current understanding. Specifically we should mention that the transient behaviour of the Langevin dynamics is apparently important, and using $\hat{V}$ instead of $V^*$ apparently matters. I intend to return to this tomorrow, but if someone takes care of it earlier, that will be just fine. -AMF}
The proposed hill climbing procedure is similar to Langevin dynamics, which is frequently used as a tool to analyze optimization algorithms or to acquire an estimate of the expected parameter values w.r.t. some posterior distribution in Bayesian learning~
\cite{welling2011sgld}.
The overdamped Langevin dynamics can be described by a stochastic differential equation (SDE) $\mathrm{d} W(t) = \nabla U(W_t) \mathrm{d}t + \sqrt{2}\mathrm{d}B_t$, where $B_t \in \RR^d$ is a $d$-dimensional Brownian motion and $U$ is a continuous differentiable function. Under some conditions, it turns out that the Langevin diffusion $(W_t)_{t\ge 0}$ converges to a unique invariant distribution $p(x) \propto \exp{(U(x))}$ \cite{chiang1987diffusionopt}. 

By apply the Euler-Maruyama discretization scheme to the SDE, we acquire the discretized version $Y_{k+1} = Y_k + \alpha_{k+1} \nabla U(Y_k) + \sqrt{2\alpha_{k+1}}Z_{k+1}$ where $(Z_k)_{k\ge 1}$ is an i.i.d. sequence of standard $d$-dimensional Gaussian random vectors and $(\alpha_k)_{k\ge 1}$ is a sequence of step sizes. This discretization scheme was used to acquire samples from the original invariant distribution $p(x) \propto \exp{(U(x))}$ through the Markov chain $(Y_k)_{k\ge 1}$ when it converges to the chain's stationary distribution \cite{roberts1996discretelangevin}. 
The distance between the limiting distribution of $(Y_k)_{k\ge 1}$ and the invariant distribution of the underlying SDE has been characterized through various bounds~\cite{durmus2017unajustedla}.


When we perform hill climbing, the parameter $\theta$ is constant at each time step $t$.
By choosing the function $U$ in the SDE above to be equal to $V_\theta$, we see that the state distribution $p(s)$ in our search-control queue is approximately\footnote{Different assumptions on $(\alpha_k)_{k\ge 1}$ and properties of $U$ can give convergence claims with different strengths. Also refer to~\cite{welling2011sgld} for the discussion on the use of a preconditioner.}
\[
\setlength{\abovedisplayskip}{3pt}
p(s) \propto \exp{(V_\theta(s))}.
\setlength{\belowdisplayskip}{3pt}
\]
An important difference between the theoretical limiting distribution and the actual distribution acquired by our hill climbing method is that our trajectories would also include the states during the \emph{burn-in} or transient period, which refers to the period before the stationary behavior is reached.
We would want to point out that those states play an essential role in improving learning efficiency as we will demonstrate in section~\ref{sec:searchcontrol-compare}.

\section{Hill Climbing Dyna}\label{sec:HC-Dyna-RL-Algorithm}
In this section, we provide the full algorithm, called Hill Climbing Dyna, summarized in Algorithm~\ref{alg_hcdyna}. The key component is to use the Hill Climbing procedure developed in the previous section, to generate states for search-control (SC). To ensure some separation between states in the search-control queue, we use a threshold $\epsilon_a$ to decide whether or not to add a state into the queue. We use a simple heuristic to set this threshold on each step, as the following sample average: $\epsilon_a \approx \epsilon_a^{(T)} = \sum_{t=1}^T \frac{||s_{t+1}-s_t||_2/\sqrt{d}}{T}$. The start state for the gradient ascent is randomly sampled from the ER buffer. 
\begin{algorithm}[t]
	\caption{HC-Dyna}
	\label{alg_hcdyna}
	\begin{algorithmic}
	        \STATE \textbf{Input:}  budget $k$ for the number of gradient ascent steps (e.g., $k = 100$), stochasticity $\eta$ for gradient ascent (e.g., $\eta = 0.1$), $\rho$ percentage of updates from SC queue (e.g., $\rho = 0.5$), $d$ the number of state variables, i.e. $\States \subset \RR^d$
		\STATE Initialize empty SC queue $\bsc$ and ER buffer $\ber$
		\STATE $\covmathat \gets \eye$  \ \ \ (empirical covariance matrix)
		\STATE $\mu_{ss} \gets \mathbf{0} \in \RR^{d\times d}, \mu_s \gets \mathbf{0} \in \RR^d$  \ \ \ (auxiliary variables for computing empirical covariance matrix, sample average will be maintained for $\mu_{ss}, \mu_s$)
		\STATE $\epsilon_a \gets 0$ \ \ \ (threshold for accepting a state) 
		\FOR {$t = 1, 2, \ldots$}
		\STATE Observe $(s, a, s', r)$ and add it to $\ber$
		\STATE $\mu_{ss} \gets \frac{ \mu_{ss}(t-1) + ss^\top}{t}, \mu_{s} \gets \frac{ \mu_{s}(t-1) + s}{t}$
		\STATE $\covmathat \gets \mu_{ss} - \mu_s \mu_s^\top$
		\STATE $\epsilon_a \gets (1-\beta) \epsilon_a + \beta \text{distance}(s', s)$ for $\beta = 0.001$
		\STATE Sample $s_0$ from $\ber$, $ \tilde{s} \gets \infty$
		\FOR {$i = 0, \ldots, k$ } 
		\STATE $g_{s_i} \gets \nabla_s V(s_i) = \nabla_s \max_a Q_\theta(s_i, a)$
		\STATE $s_{i+1}\! \gets \Pi(s_i + \frac{0.1}{||\hat{\Sigmamat}_s g_{s_i}||} \hat{\Sigmamat}_s g_{s_i} + X_i), X_i \! \sim \mathcal{N}(0, \eta \hat{\Sigmamat}_\svec)$
		\IF {distance$(\tilde{s}, s_{i+1}) \ge \epsilon_a$}
		\STATE Add $s_{i+1}$ into $\bsc$, $\tilde{s} \gets s_{i+1}$
		\ENDIF		
		\ENDFOR
		\FOR {$n$ times}
		\STATE Sample a mixed mini-batch $b$, with proportion $\rho$ from $\bsc$ and $1-\rho$ from $\ber$
		\STATE Update parameters $\theta$ (i.e. DQN update) with $b$
		\ENDFOR
		\ENDFOR
	\end{algorithmic}
\end{algorithm}

In addition to using this new method for search control, 
we also found it beneficial to include updates on the experience generated in the real world. The mini-batch sampled for training has $\rho$ proportion of transitions generated by states from the SC queue, and $1-\rho$ from the ER buffer. For example, for $\rho = 0.75$ with a mini-batch size of $32$, the updates consists of $24(=32\times 0.75)$ transitions generated from states in the SC queue and 6 transitions from the ER buffer. Previous work using Dyna for learning NN value functions also used such mixed mini-batches \cite{holland2018dynaplanshape}. 

One potential reason this addition is beneficial is that it alleviates issues with heavily skewing the sampling distribution to be off-policy. Tabular Q-learning is an off-policy learning algorithm, which has strong convergence guarantees under mild assumptions \cite{Tsitsiklis1994}. When moving to function approximation, however, convergence of Q-learning is much less well understood.
\todo{DQN is not really Q-learning; it is a Fitted Q-Iteration algorithm, which has been studied by Munos, Szepesvari, and myself. -AMF}
The change in sampling distribution for the states could significantly impact convergence rates, and potentially even cause divergence. Empirically, previous prioritized ER work pointed out that skewing the sampling distribution from the ER buffer can lead to a biased solution~\cite{schaul2016prioritized}. Though the ER buffer is not on-policy, because the policy is continually changing, the distribution of states is closer to the states that would be sampled by the current policy than those in SC. \todo{Is this what was intended to be said? -AMF}
Using mixed states from the ER buffer, and those generated by Hill Climbing, could alleviate some of the issues with this skewness. 

Another possible reason that such mixed sampling could be necessary is due to model error. The use of real experience could mitigate issues with such error. We found, however, that this mixing has an effect even when using the true model. This suggests that this phenomenon indeed is related to the distribution over states.  

\begin{figure}
	\subfigure[(Continuous state) GridWorld]{
		\includegraphics[width=\figwidthfour]{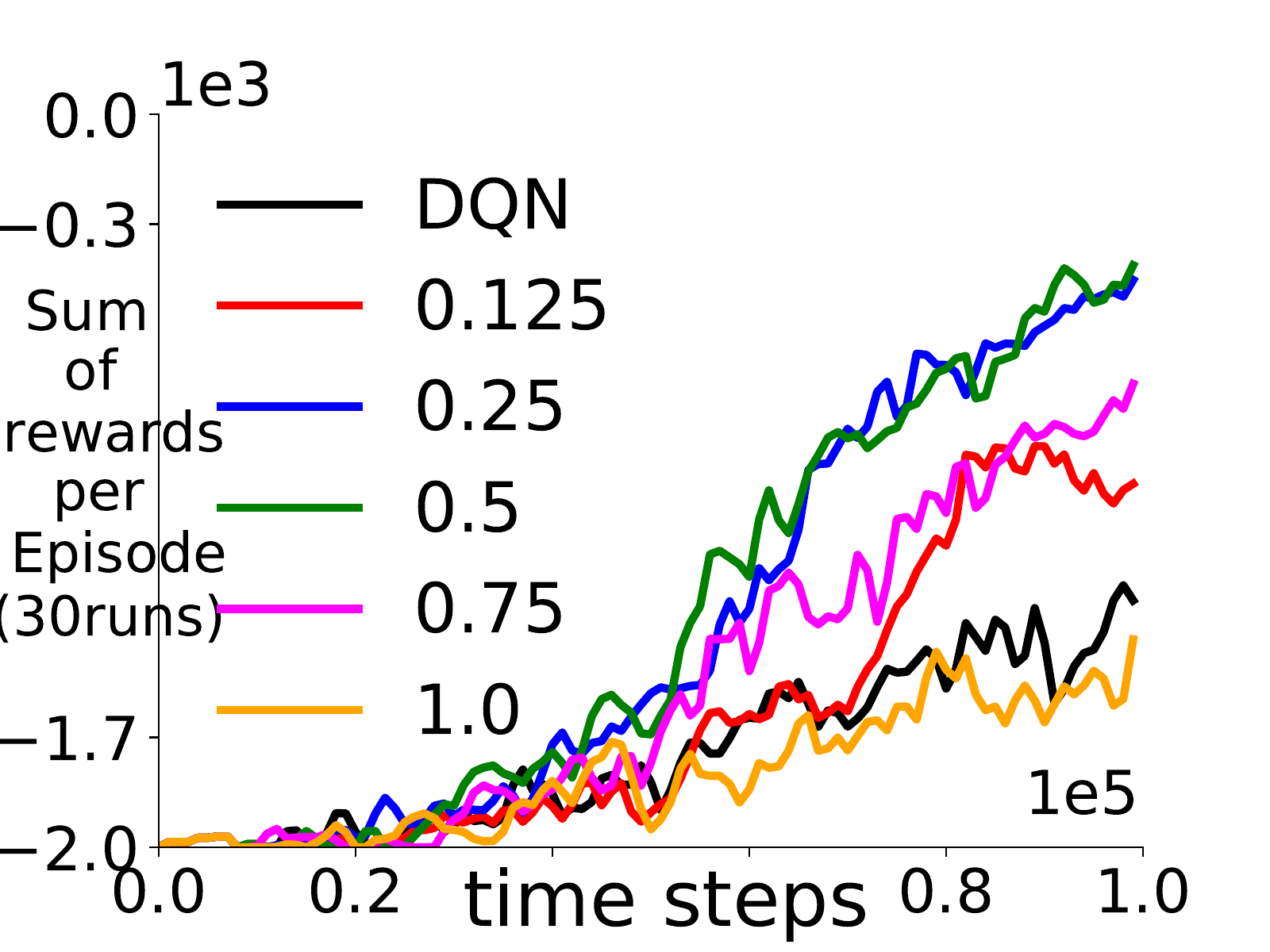}}\label{fig:mixcontigd}\hfil
	\subfigure[TabularGridWorld]{
		\includegraphics[width=\figwidthfour]{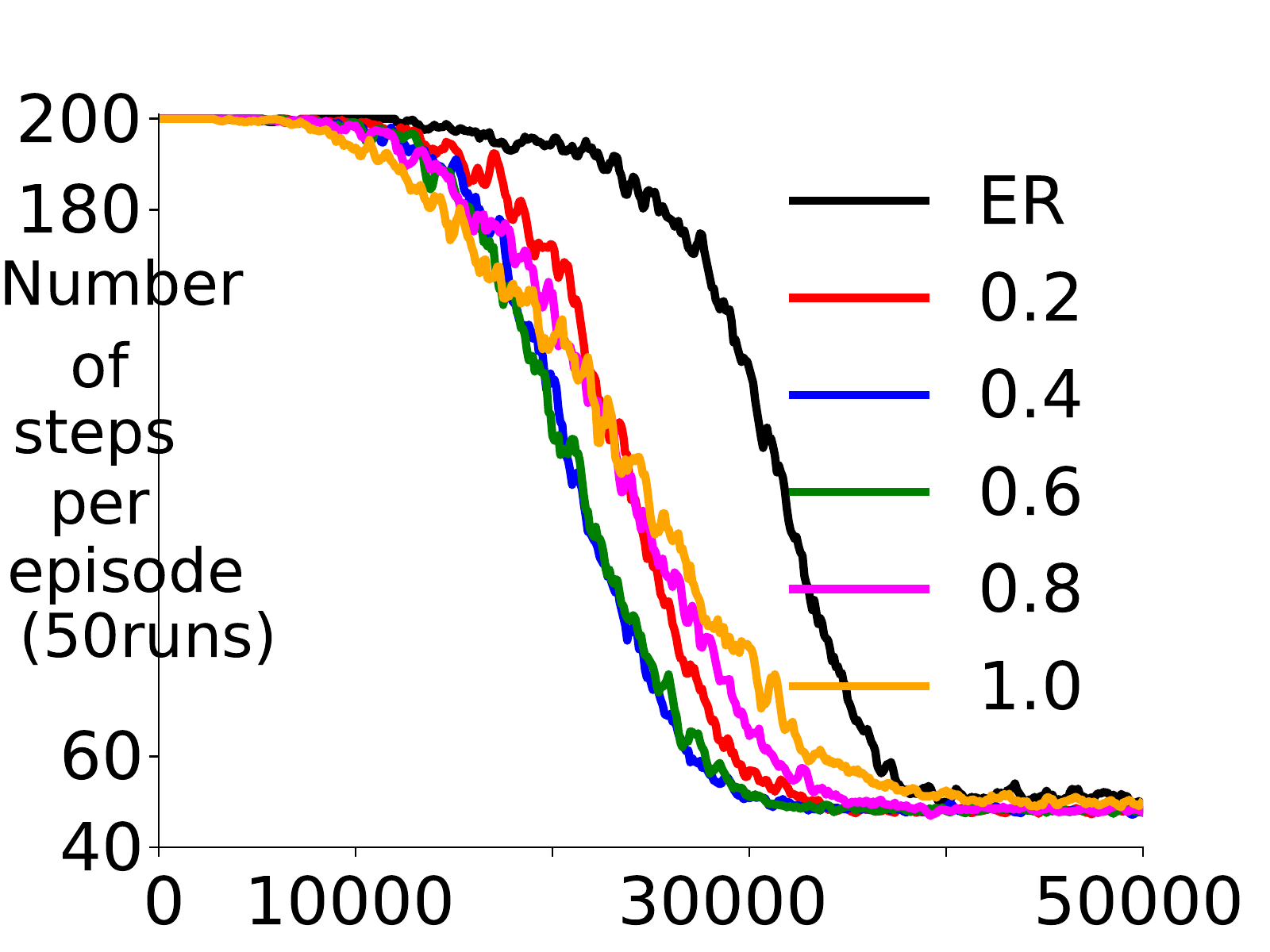}} \label{fig:mixtabulargd}
	\vspace{-0.25cm}
	\caption{
		The effect of mixing rate on learning performance. The numerical label means HC-Dyna with a certain mixing rate.
	}\label{fig:mixingrate}
		\vspace{-0.25cm}
\end{figure}

We provide a small experiment in the GridWorld, depicted in Figure \ref{fig:gridworldexample}, using both a continuous-state and a discrete-state version. We include a discrete state version, so we can demonstrate that the effect persists even in a tabular setting when Q-learning is known to be stable. The continuous-state setting uses NNs---as described more fully in Section \ref{sec:HC-Dyna-Experiments}---with a mini-batch size of 32. For the tabular setting, the mini-batch size is 1; updates are randomly selected to be from the SC queue or ER buffer proportional to $\rho$. Figure~\ref{fig:mixingrate} shows the performance of HC-Dyna as the mixing proportion increases from $0$ (ER only) to $1.0$ (SC only).
In both cases, a mixing rate around $\rho = 0.5$ provides the best results. Generally, using too few search-control samples do not improve performance; focusing too many updates on search-control samples seems to slightly speed up early learning, but then later learning suffers. In all further experiments in this paper, we set $\rho = 0.5$.

\section{Experiments}\label{sec:HC-Dyna-Experiments}

In this section, we demonstrate the utility of (DQN-)HC-Dyna in several benchmark domains, and then analyze the learning effect of different sampling distributions to generate states for the search-control queue. 

\subsection{Results in Benchmark Domains}\label{sec:classicaldomains-results}

In this section, we present empirical results on four classic domains: the GridWorld (Figure~\ref{fig:gridworld}), MountainCar, CartPole and Acrobot. We present both discrete and continuous action results in the GridWorld, and compare to DQN for the discrete control and to Deep Deterministic Policy Gradient (DDPG) for the continuous control~\cite{tim2016ddpg}. The agents all use a two-layer NN, with ReLU activations and 32 nodes in each layer. We include results using both the true model and the learned model, on the same plots. We further include multiple planning steps $n$, where for each real environment step, the agent does $n$ updates with a mini-batch of size 32. 

In addition to ER, we add an on-policy baseline called \textbf{OnPolicy-Dyna}. This algorithm samples a mini-batch of states (not the full transition) from the ER buffer, but then generates the next state and reward using an on-policy action. This baseline distinguishes when the gain of HC-Dyna algorithm is due to on-policy sampled actions, rather than because of the states in our search-control queue.

\begin{figure}[t]
		\subfigure[plan steps 1]{
			\includegraphics[width=\figwidthsix]{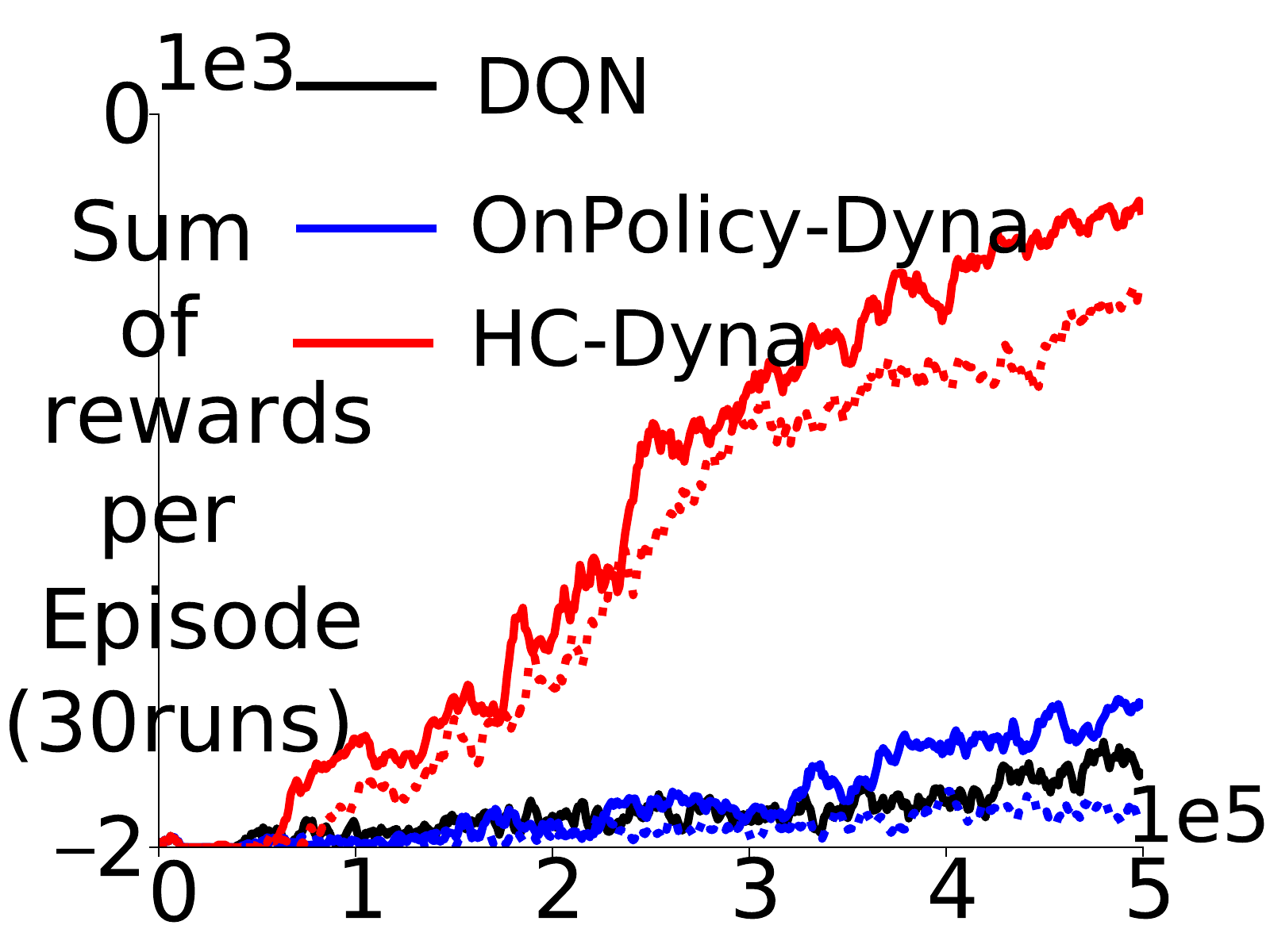}}
	\subfigure[plan steps 10]{
		\includegraphics[width=\figwidthsix]{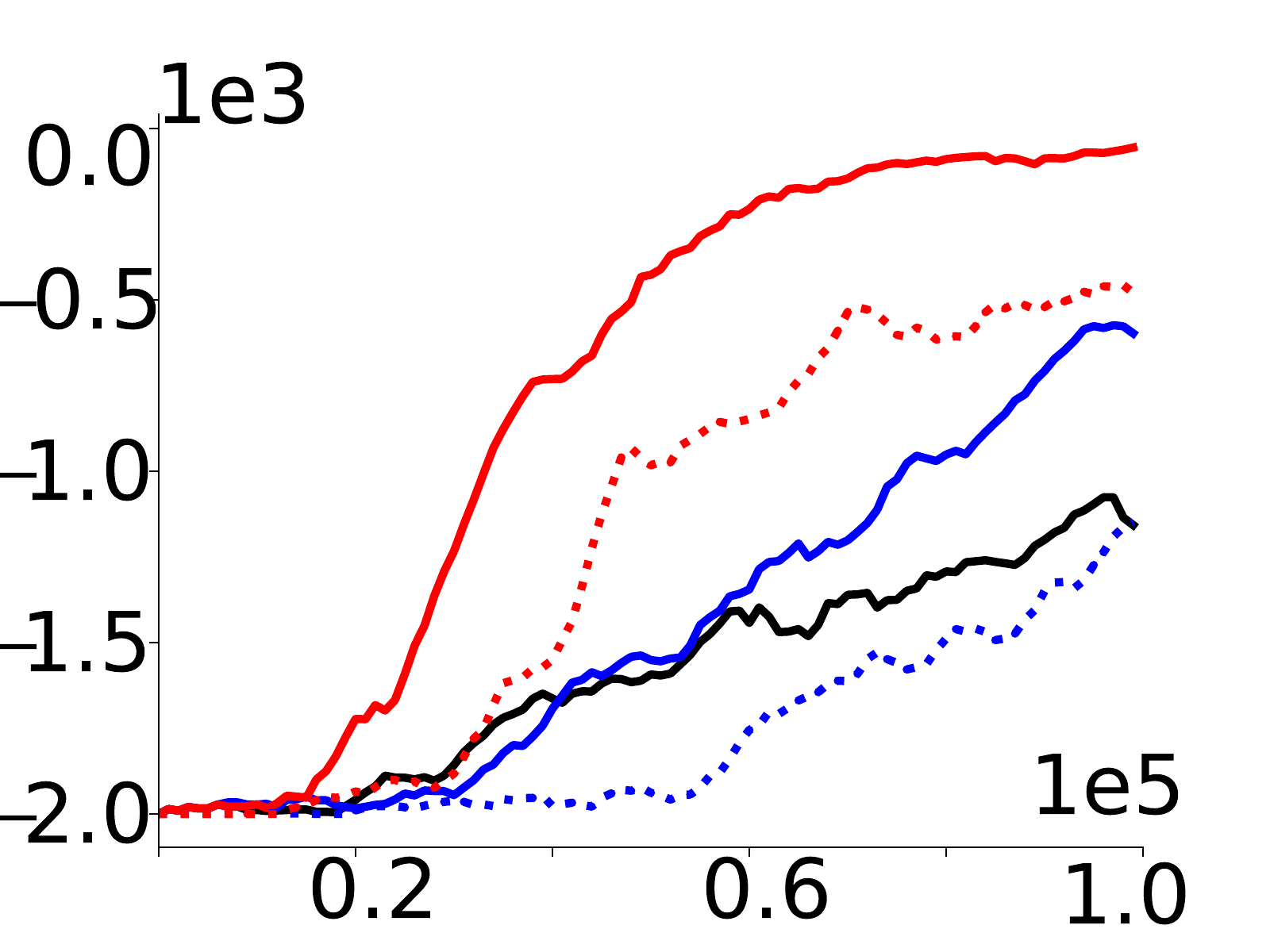}}
	\subfigure[plan steps 30]{
		\includegraphics[width=\figwidthsix]{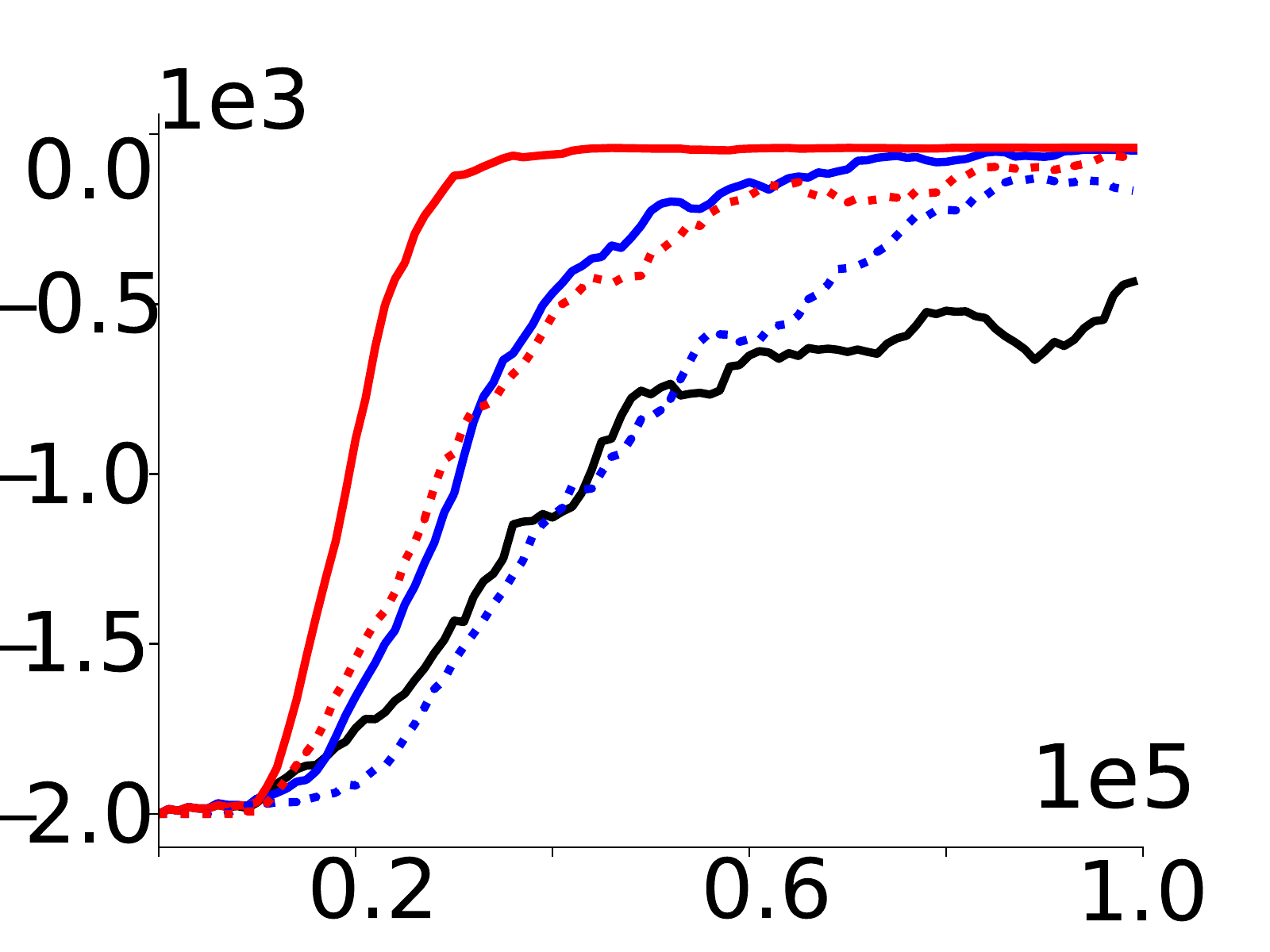}}
	\subfigure[plan steps 1]{
		\includegraphics[width=\figwidthsix]{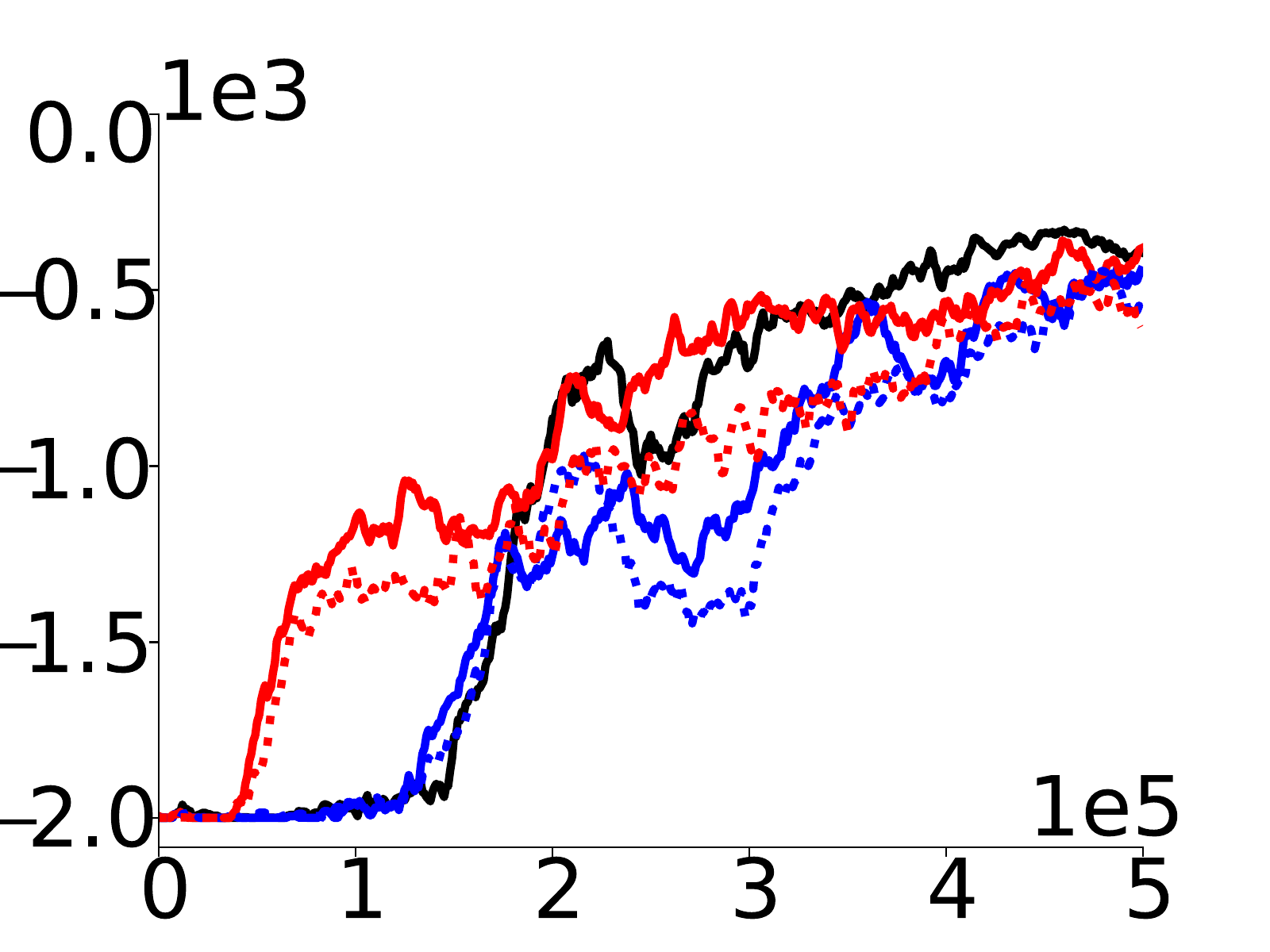}}
	\subfigure[plan steps 10]{
		\includegraphics[width=\figwidthsix]{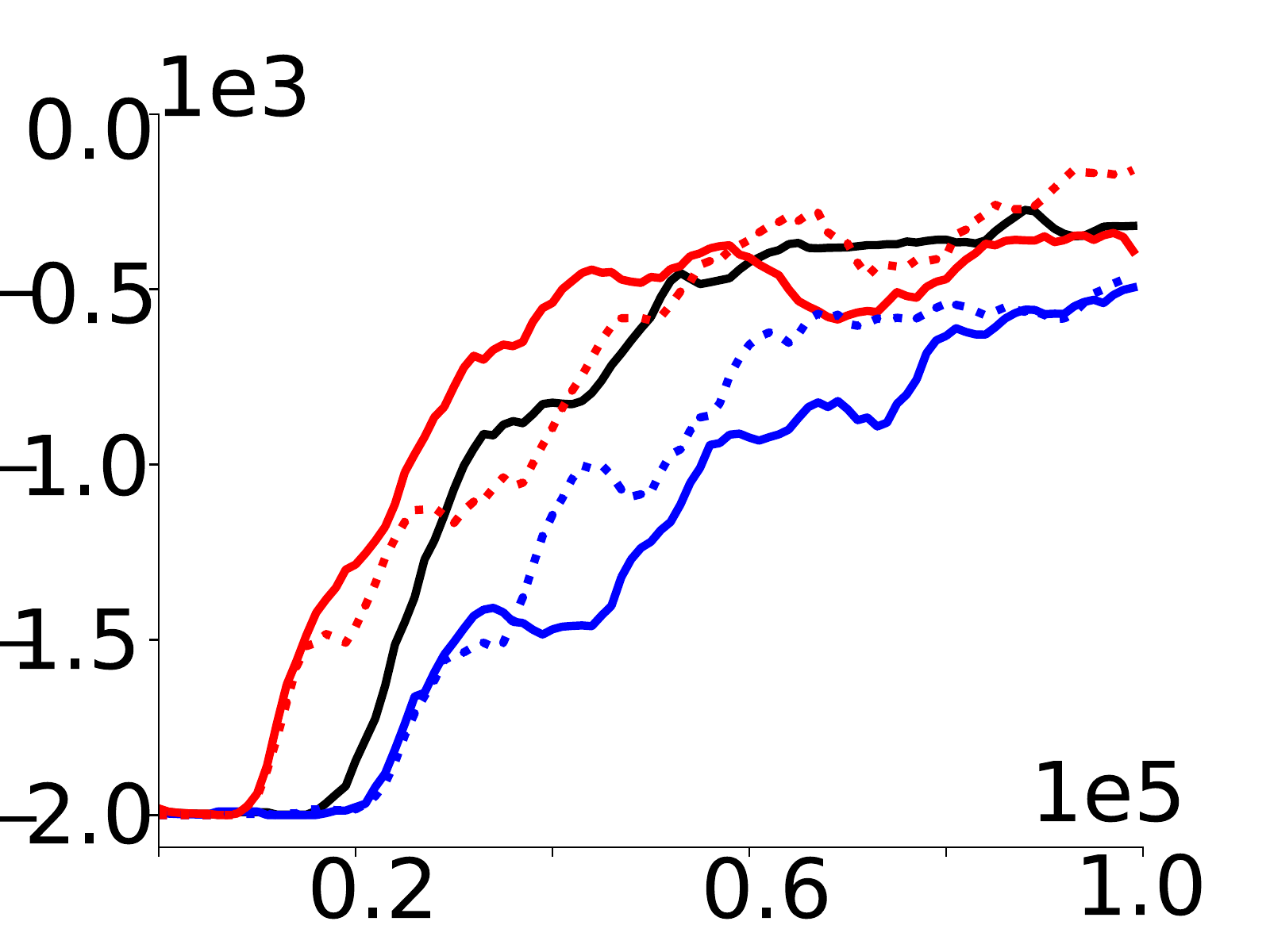}}
	\subfigure[plan steps 30]{
		\includegraphics[width=\figwidthsix]{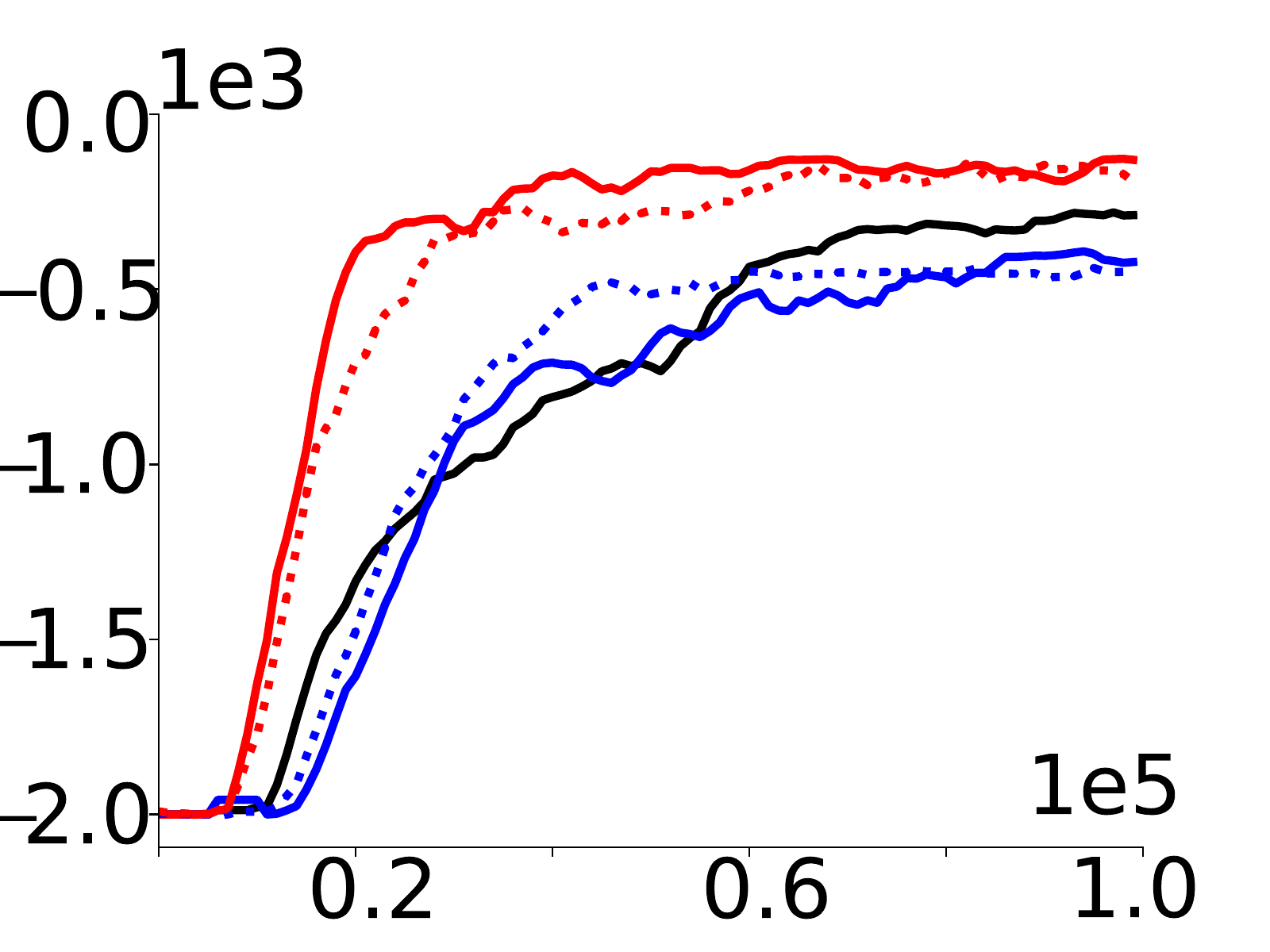}}
		\subfigure[plan steps 1]{
			\includegraphics[width=\figwidthsix]{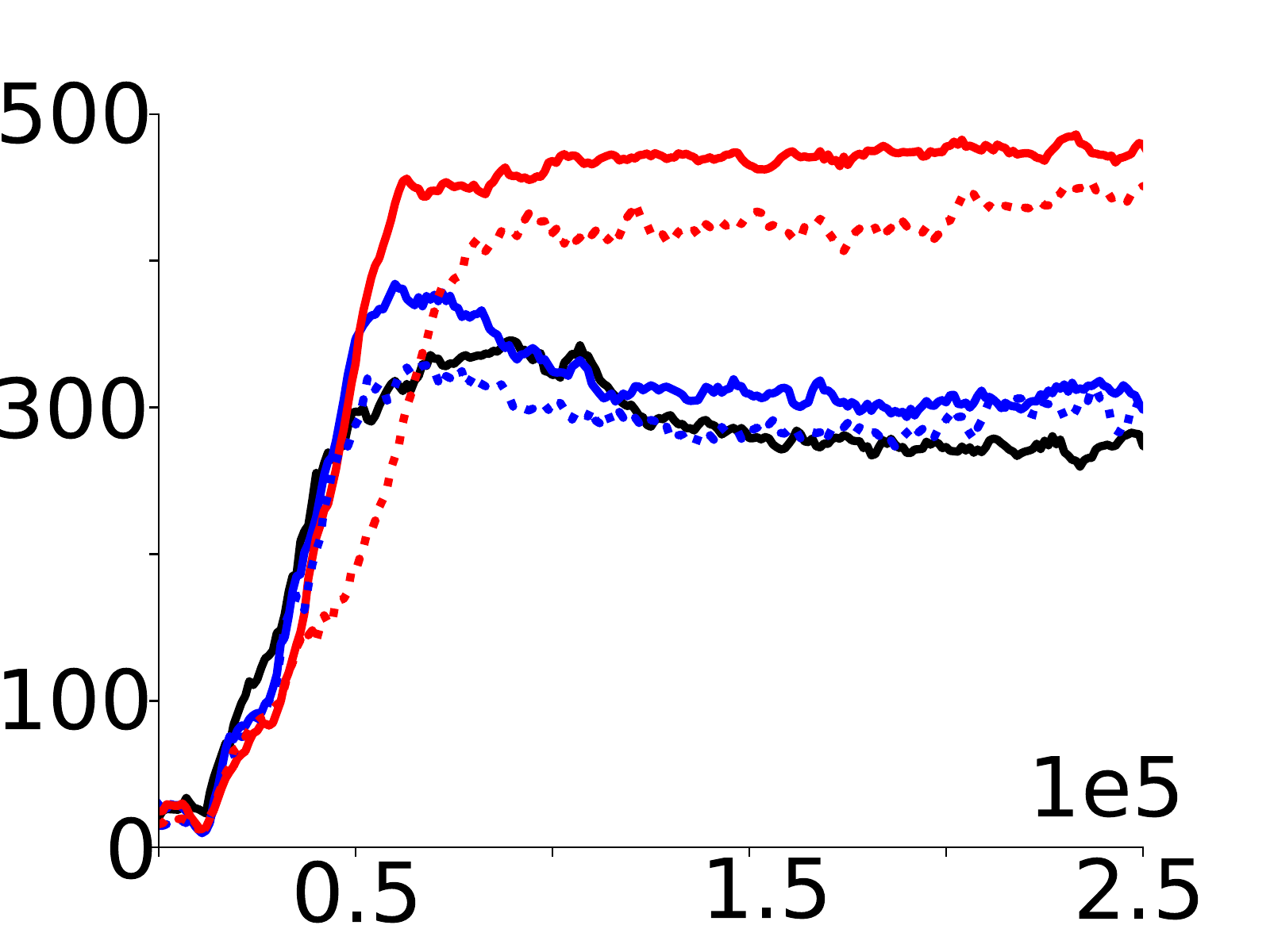}}
		\subfigure[plan steps 10]{
			\includegraphics[width=\figwidthsix]{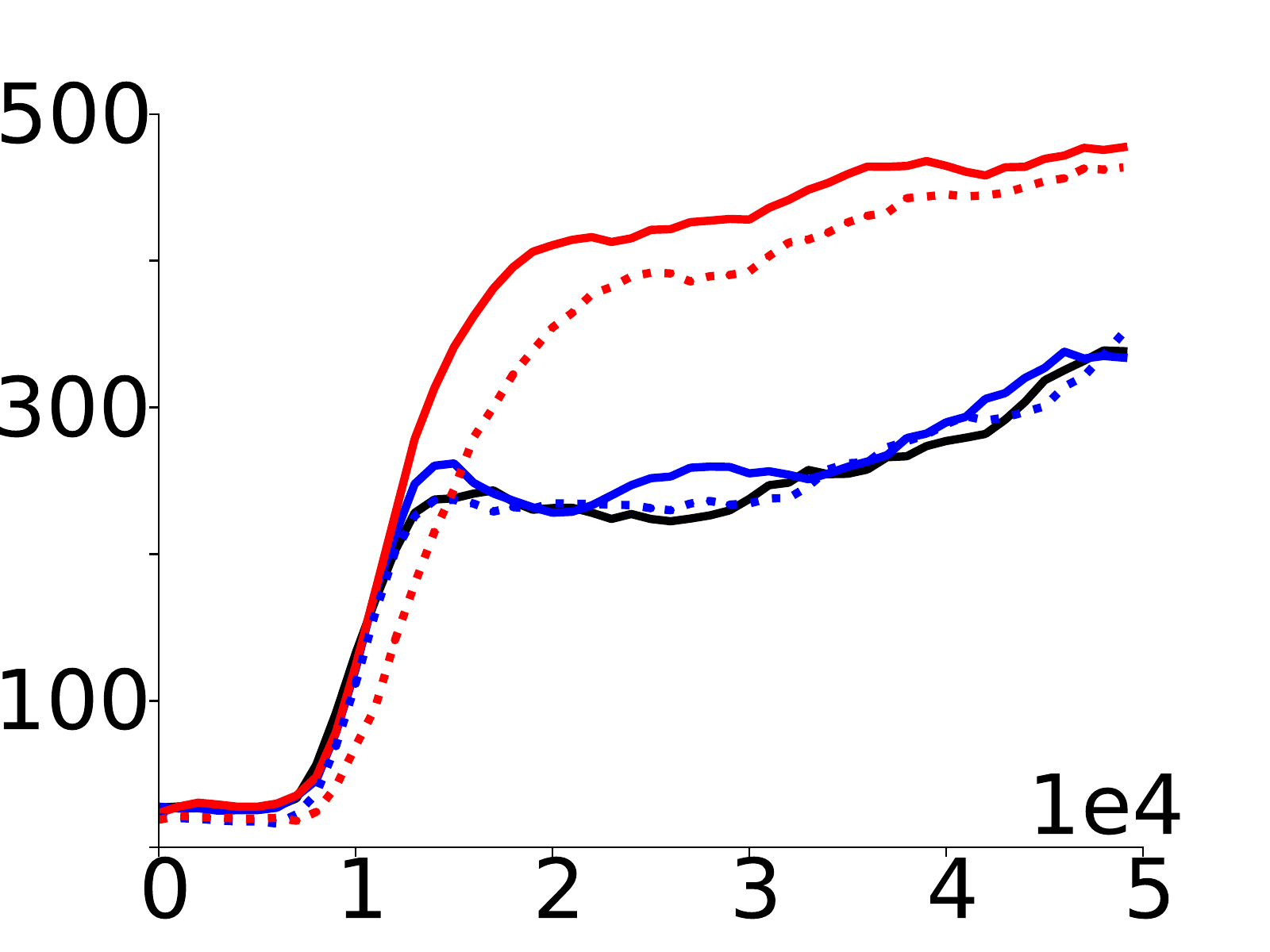}}
		\subfigure[plan steps 30]{
			\includegraphics[width=\figwidthsix]{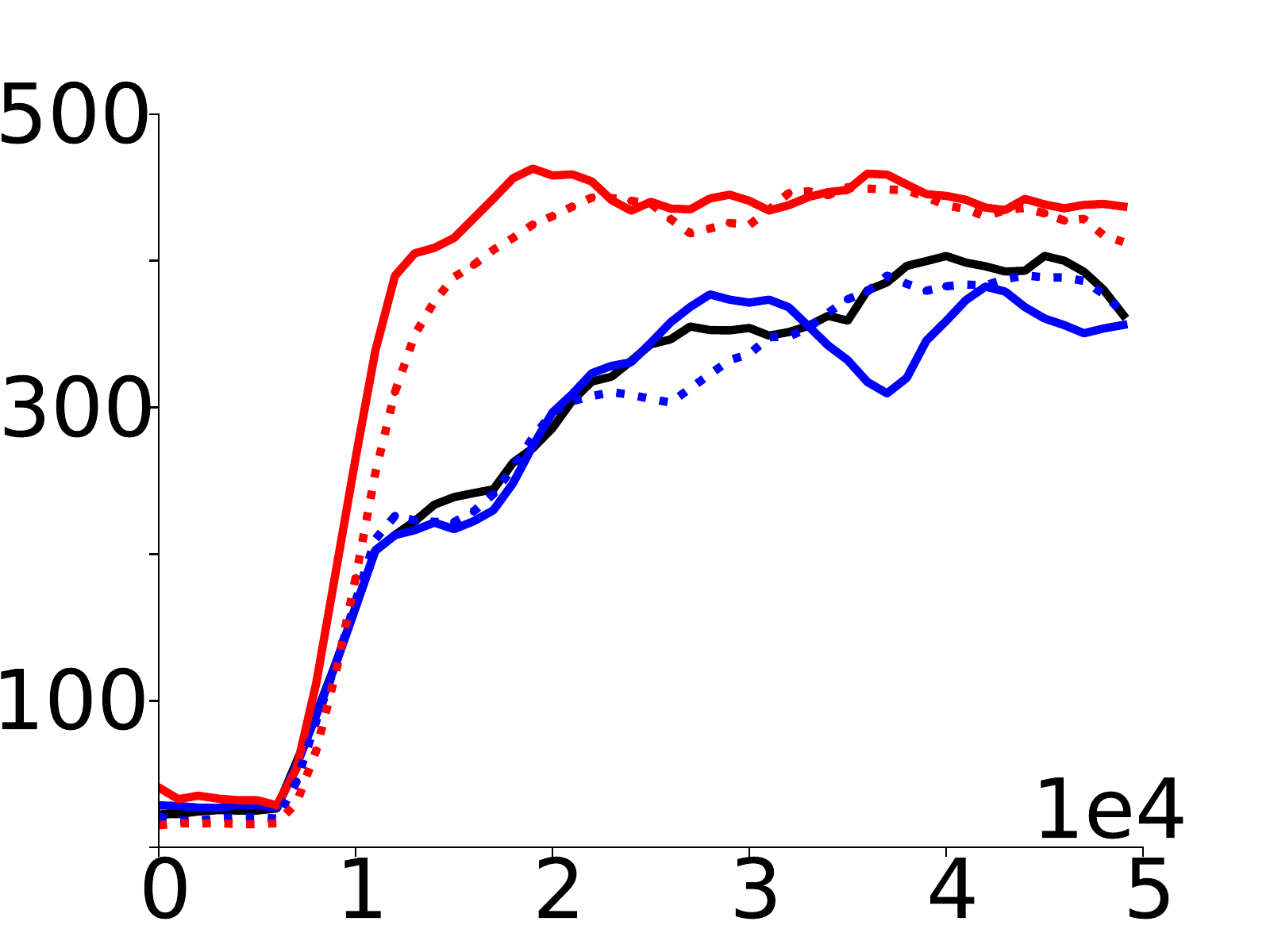}}
		\subfigure[plan steps 1]{
			\includegraphics[width=\figwidthsix]{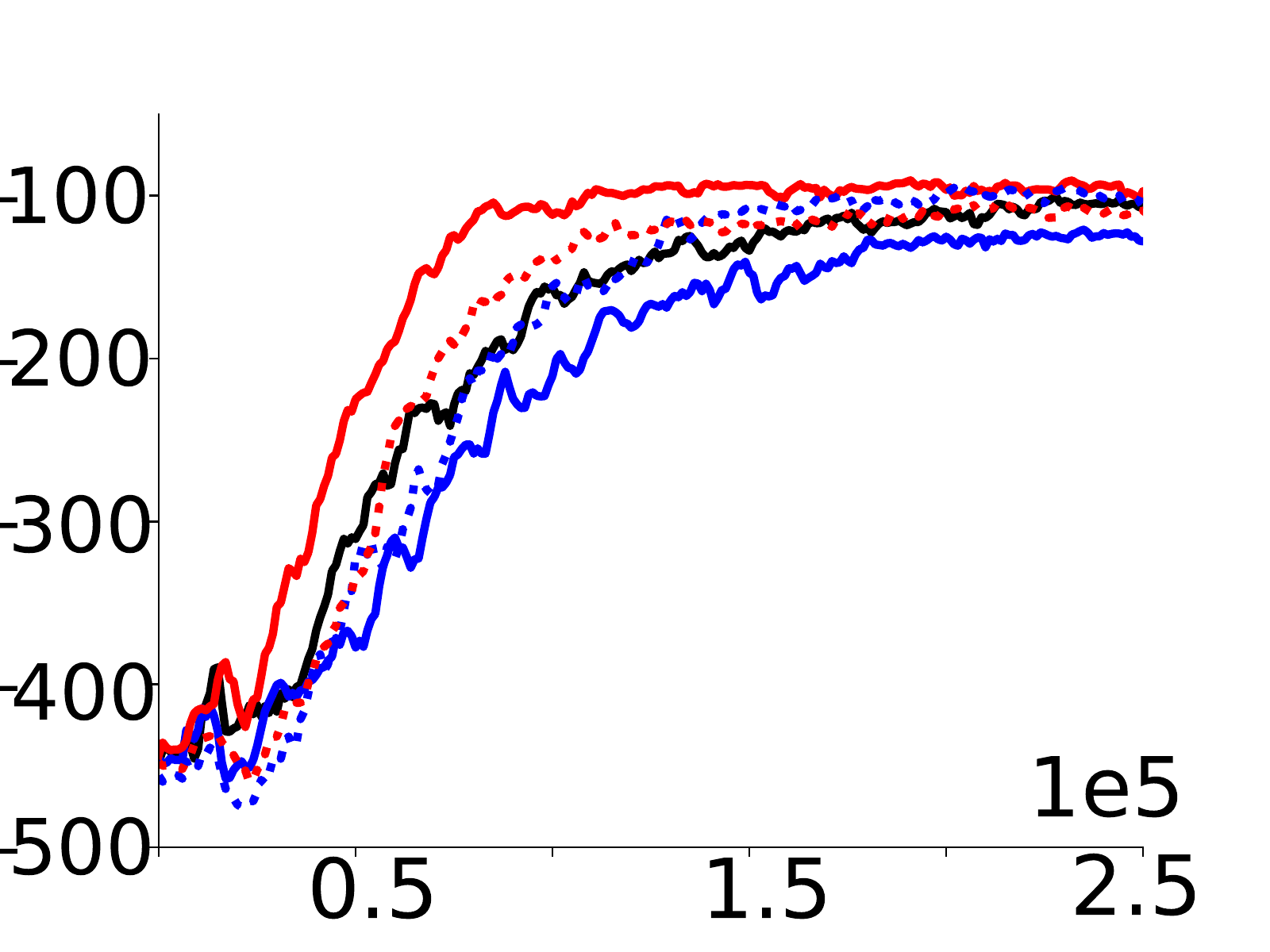}}
		\subfigure[plan steps 10]{
			\includegraphics[width=\figwidthsix]{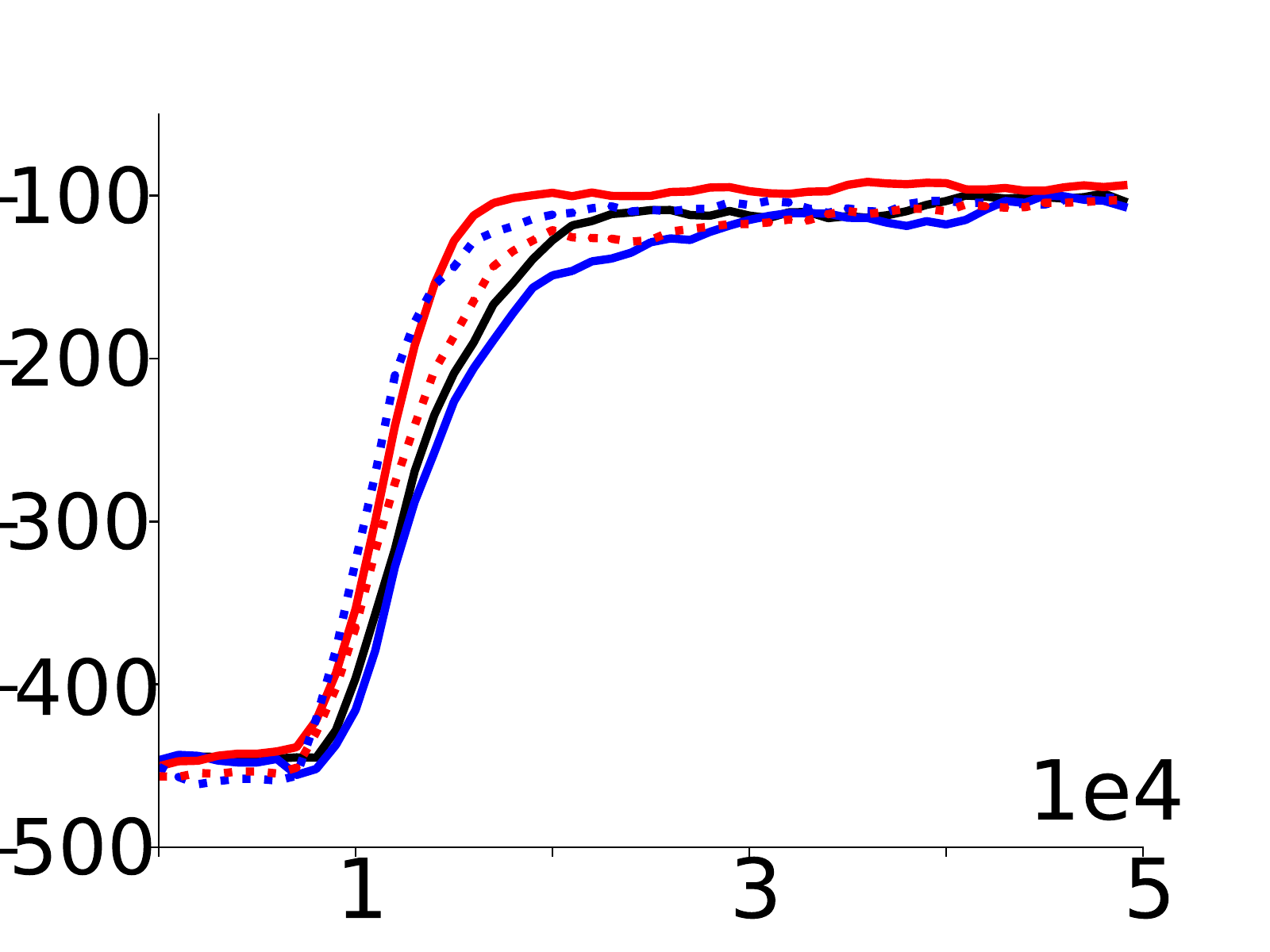}}
		\subfigure[plan steps 30]{
			\includegraphics[width=\figwidthsix]{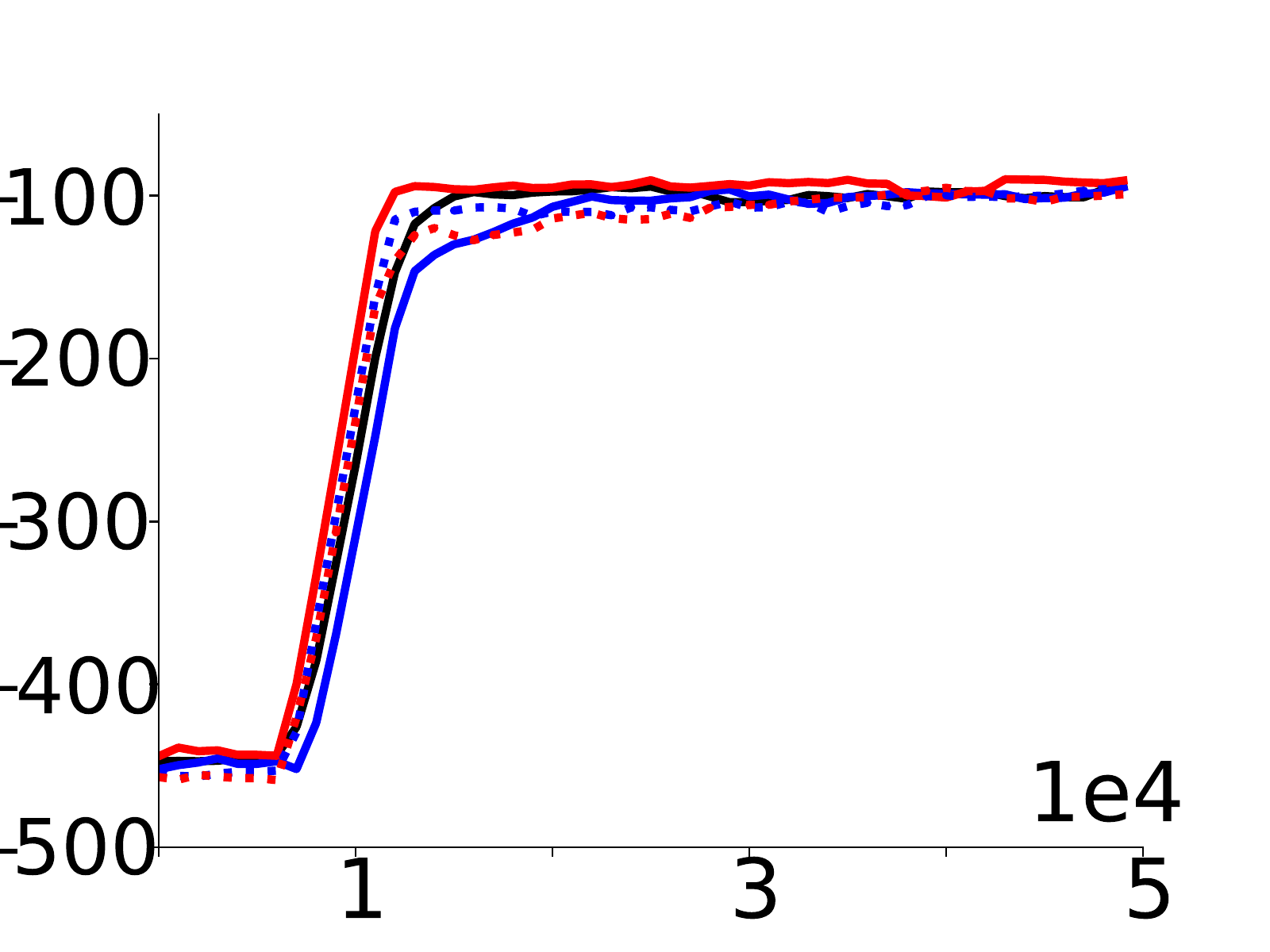}}
	\caption{
		Evaluation curves (sum of episodic reward v.s. environment time steps) of {\color{red} (DQN-)HC-Dyna}, {\color{blue} (DQN-)OnPolicy-Dyna}, {\color{black} DQN}
		on GridWorld (a-c), MountainCar-v0 (d-f), CartPole-v1 (g-i) and Acrobot-v1 (j-l). The curves plotted by \emph{dotted lines} are using online learned models. Results are averaged over $30$ random seeds. 
	}\label{fig:experiments}
\end{figure}

\subsubsection{Discrete Action}
The results in Figure~\ref{fig:experiments} show that (a) HC-Dyna never harms performance over ER and OnPolicy-Dyna, and in some cases significantly improves performance, (b) these gains persist even under learned models and (c) there are clear gains from HC-Dyna even with a small number of planning steps. 
Interestingly, using multiple mini-batch updates per time step can significantly improve the performance of all the algorithms. 
DQN, however, has very limited gain when moving from $10$ to $30$ planning steps on all domains except GridWorld, whereas HC-Dyna seems to more noticeably improve from more planning steps.  This implies a possible limit of the usefulness of only using samples in the ER buffer. 

We observe that the on-policy actions does not always help.
The GridWorld domain is in fact the only one where on-policy actions (OnPolicy-Dyna) shows an advantage as the number of planning steps increase. This result provides evidence that the gain of our algorithm is due to the states in our search-control queue, rather than on-policy sampled actions.
We also see that even though both model-based methods perform worse when the model has to be learned compared to when the true model is available, HC-Dyna is consistently better than OnPolicy-Dyna across all domains/settings. 

\begin{figure}
	\subfigure[DQN]{
		\includegraphics[width=0.23\textwidth]{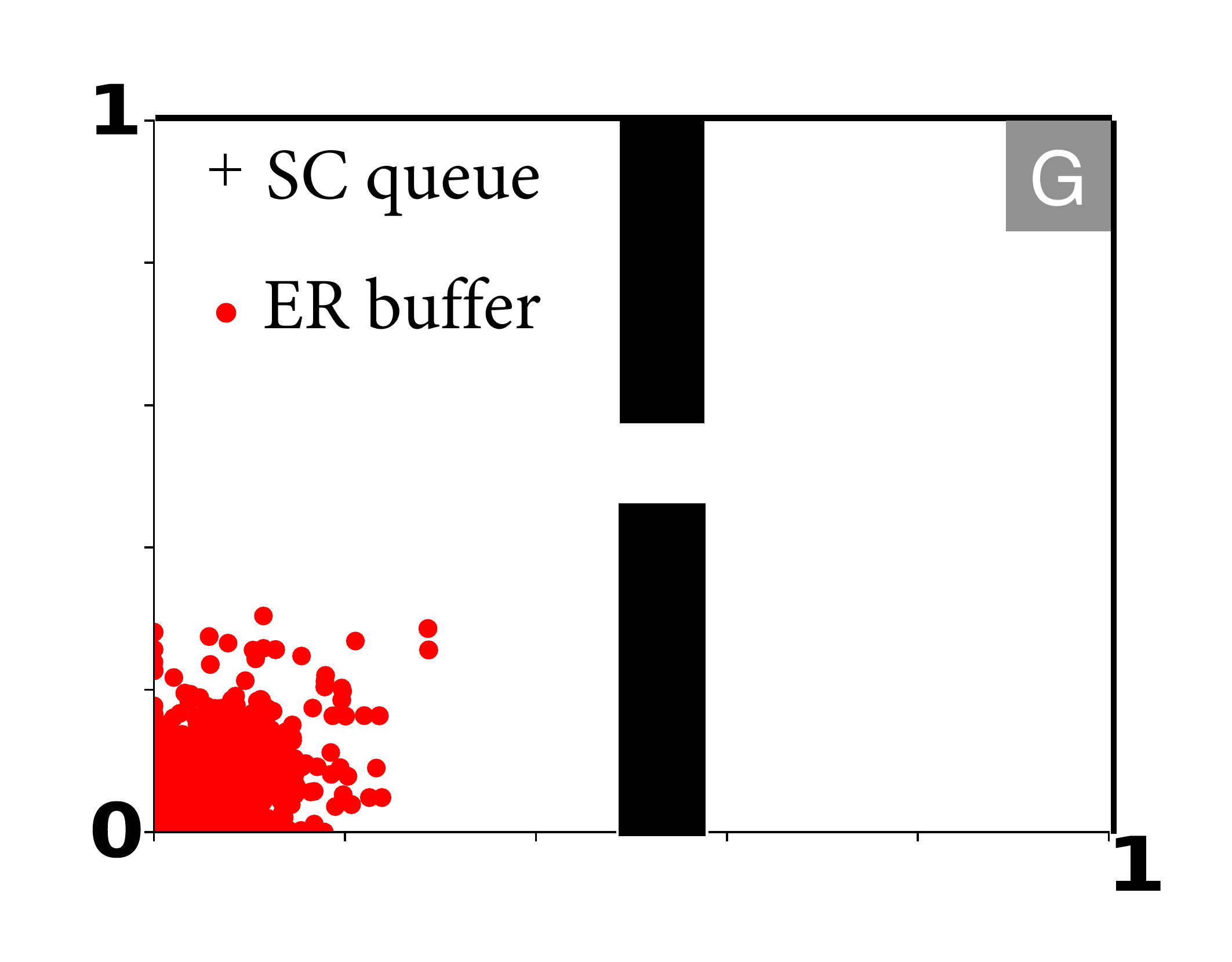} \label{fig:dqner}}\hfil
	\subfigure[HC-Dyna]{
		\includegraphics[width=0.23\textwidth]{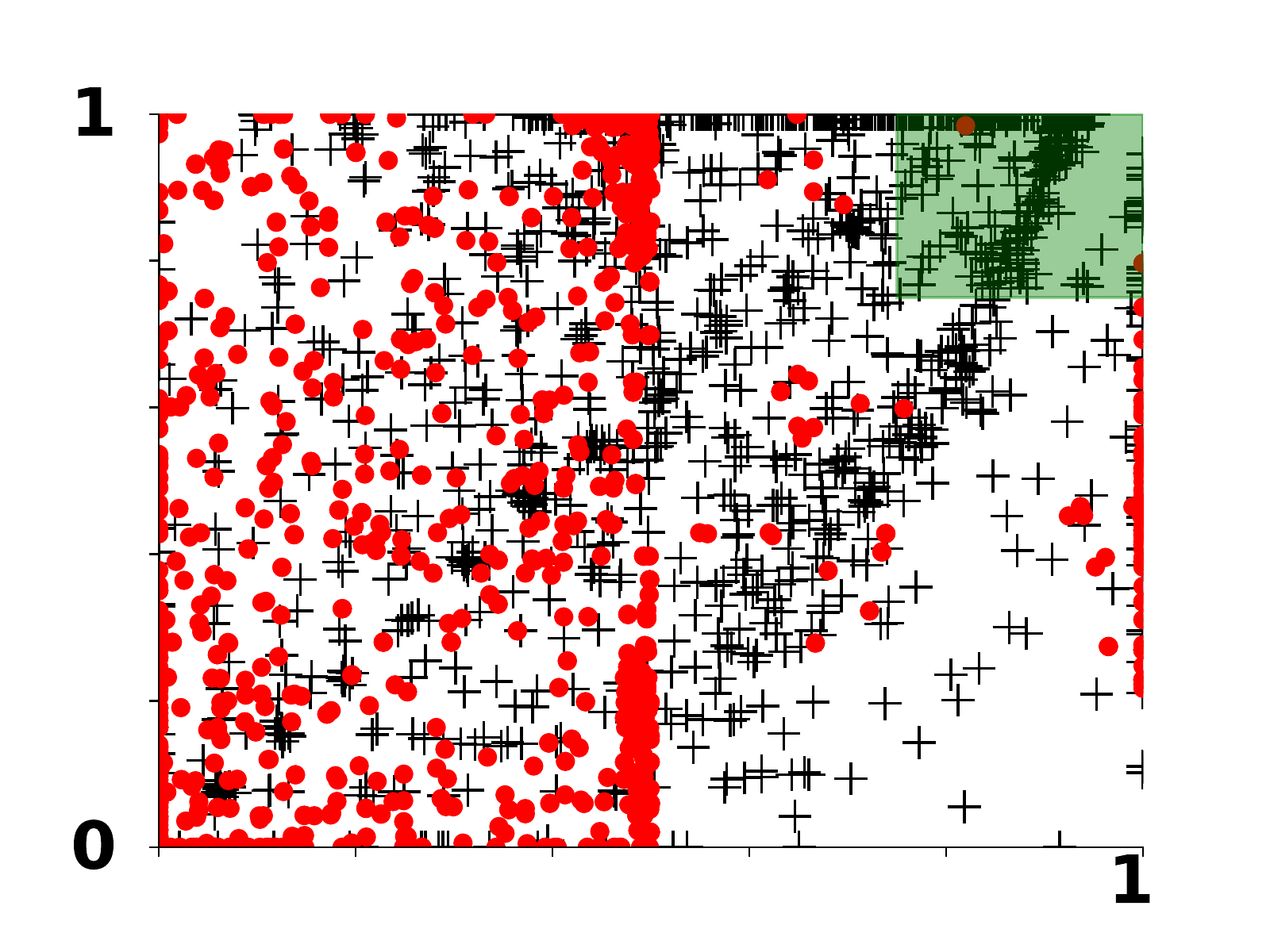} \label{fig:hcdynaerqueue}}
	\vspace{-0.35cm}
	\caption{
		Figure (a)(b) show buffer({\color{red}red $\boldsymbol{\cdot}$})/queue(black $+$) distribution on GridWorld ($\svec \in [0, 1]^2$) by uniformly sampling $2$k states. 
		(a) is showing ER buffer when running DQN, hence there is no ``$+$'' in it. (b) shows $0.2\%$ of the ER samples fall in the green shadow (i.e. high value region), while $27.8\%$ samples from the SC queue are there.
	}\label{fig:algodesign}
	\vspace{-0.25cm}
\end{figure}

To gain intuition for why our algorithm achieves superior performance, we visualize the states in the search-control queue for HC-Dyna in the GridWorld domain (Figure \ref{fig:algodesign}).
We also show the states in the ER buffer at the same time step, for both HC-Dyna and DQN to contrast. There are two interesting outcomes from this visualization. First, the modification to search-control significantly changes where the agent explores, as evidenced by the ER buffer distribution. Second, HC-Dyna has many states in the SC queue that are near the goal region even when its ER buffer samples concentrate on the left part on the square. The agent can still update around the goal region even when it is physically in the left part of the domain. 

\subsubsection{Continuous Control} 
\begin{figure}
	\includegraphics[width=0.26\textwidth]{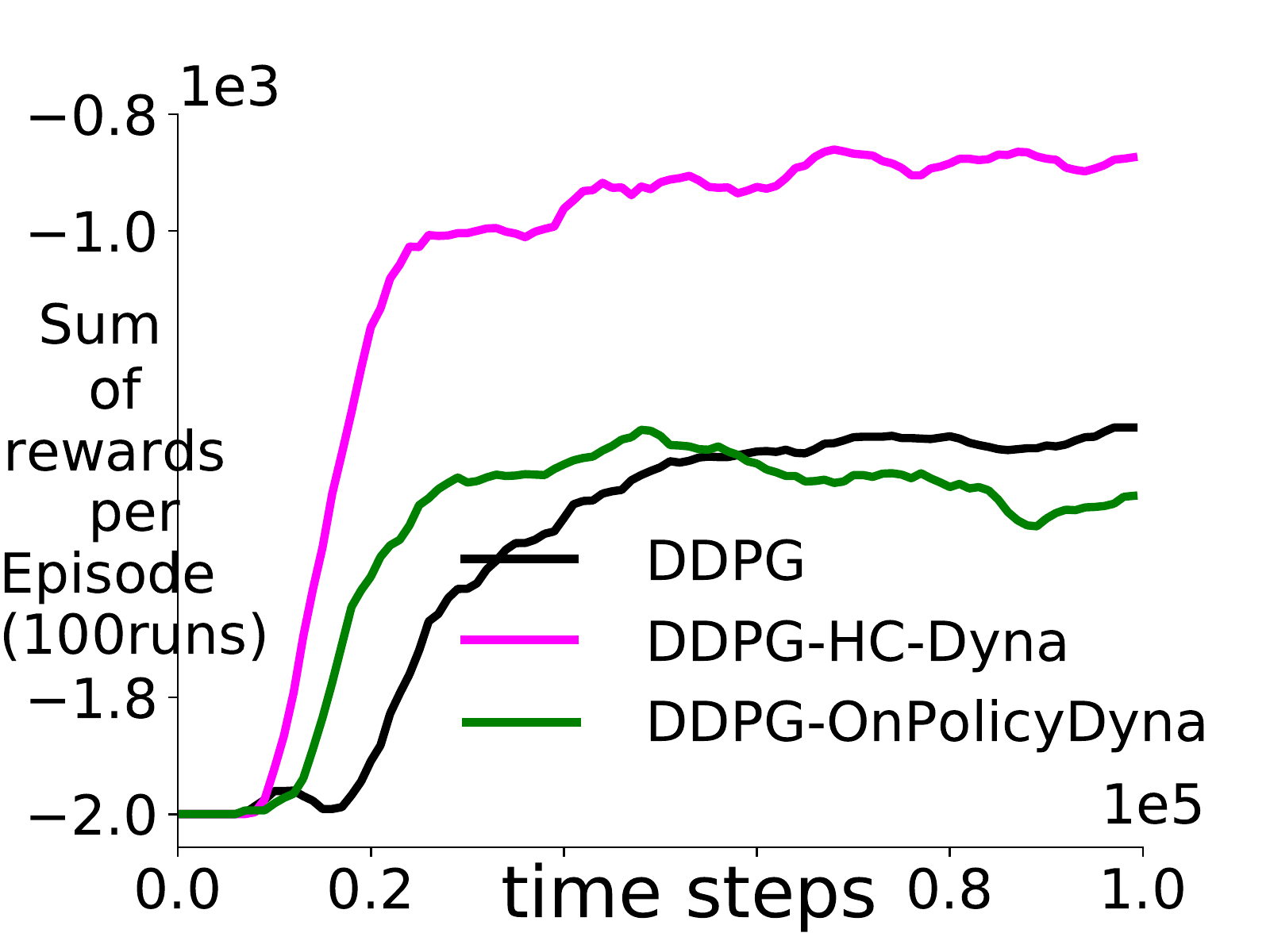}
			\caption{
				HC-Dyna for continuous control with DDPG. 
				We used $5$ planning steps in this experiment. 
			} \label{fig:contigd}
	\vspace{-0.25cm}
\end{figure}

Our architecture can easily be used with continuous actions, as long as the algorithm estimates values. 
We use DDPG \cite{tim2016ddpg} as an example for use inside HC-Dyna. DDPG is an actor-critic algorithm that uses the deterministic policy gradient theorem \cite{silver2014dpg}. Let $\pi_\psi(\cdot) : \States \rightarrow \Actions$ be the actor network parameterized by $\psi$, and $Q_\theta(\cdot, \cdot) : \States \times \Actions \rightarrow \mathbb{R}$ be the critic. Given an initial state value $s$, the gradient ascent direction can be computed by $\nabla_s Q_\psi(s, \pi_\theta(s))$. In fact, because the gradient step causes small changes, we can further approximate this gradient more efficiently using $\nabla_s Q_\psi(s, a^\ast), a^\ast \defeq \pi_\theta(s)$, without backpropagating the gradient through the actor network. We modified the GridWorld in Figure~\ref{fig:gridworld} to have action space $\Actions = [-1,1]^2$ and an action $a_t \in \Actions$ is executed as $s_{t+1} \gets s_t + 0.05a_t$. Figure~\ref{fig:contigd} shows the learning curve of DDPG, and DDPG with OnPolicy-Dyna and with HC-Dyna. As before, HC-Dyna shows significant early learning benefits and also reaches a better solution. This highlights that improved search-control could be particularly effective for algorithms that are known to be prone to local minima, like Actor-Critic algorithms.   

\subsection{Investigating Sampling Distributions for Search-control}\label{sec:searchcontrol-compare}
We next investigate the importance of two choices in HC-Dyna: (a) using trajectories to high-value regions and (b) using the agent's value estimates to identify these important regions. To test this, we include following sampling methods for comparison: (a) HC-Dyna: hill climbing by using $\hat{V}_{\theta}$ (our algorithm); (b) Gibbs: sampling $\propto \exp{(\hat{V}_{\theta})}$; (c) HC-Dyna-Vstar: hill climbing by using $V^*$ and (d) Gibbs-Vstar: sampling $\propto \exp{(V^*)}$, where $V^*$ is a pre-learned optimal value function. We also include the baselines OnPolicyDyna, ER and Uniform-Dyna, which uniformly samples states from the whole state space. All strategies mix with ER, using $\rho = 0.5$, to better give insight into performance differences. 


To facilitate sampling from the Gibbs distribution and computing the optimal value function, we test on a simplified TabularGridWorld domain of size $20\times 20$, without any obstacles. Each state is represented by an integer $i \in \{1, ..., 400\}$, assigned from bottom to top, left to right on the square with $20\times 20$  grids. 
HC-Dyna and HC-Dyna-Vstar assume that the state space is continuous on the square $[0, 1]^2$ and each grid can be represented by its center's $(x, y)$ coordinates. We use the finite difference method for hill climbing. 

\subsubsection{Comparing to the Gibbs Distribution}
\begin{figure}[t]
 \vspace{-0.2cm}
	\centering 
		\subfigure[TabularGridWorld]{
			\includegraphics[width=\figwidthfour]{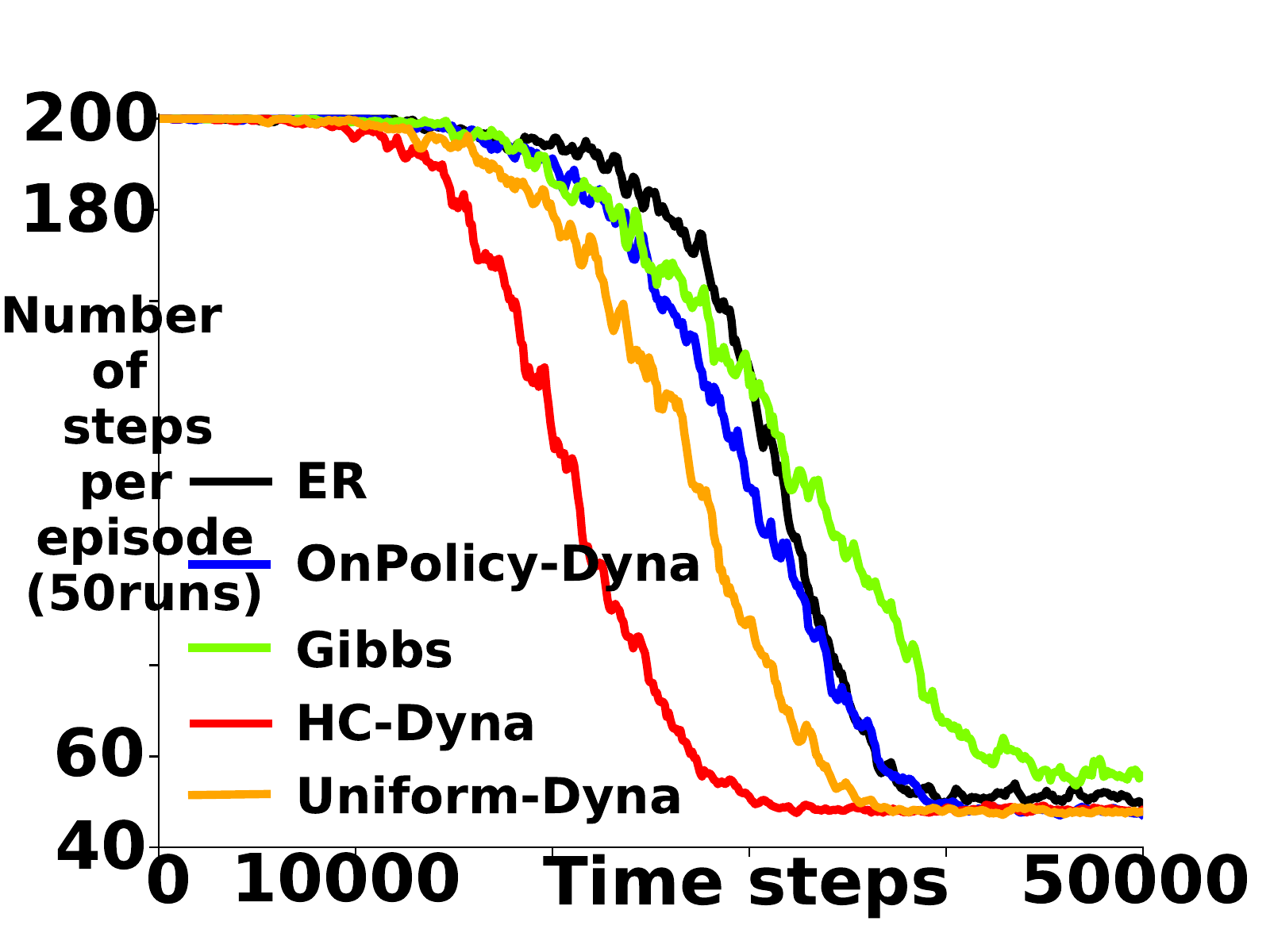}\hfil
			 \label{fig:tabulargd_compare_scqueue}}
		\subfigure[HC-Dyna vs. Gibbs]{
			\includegraphics[width=\figwidthfour]{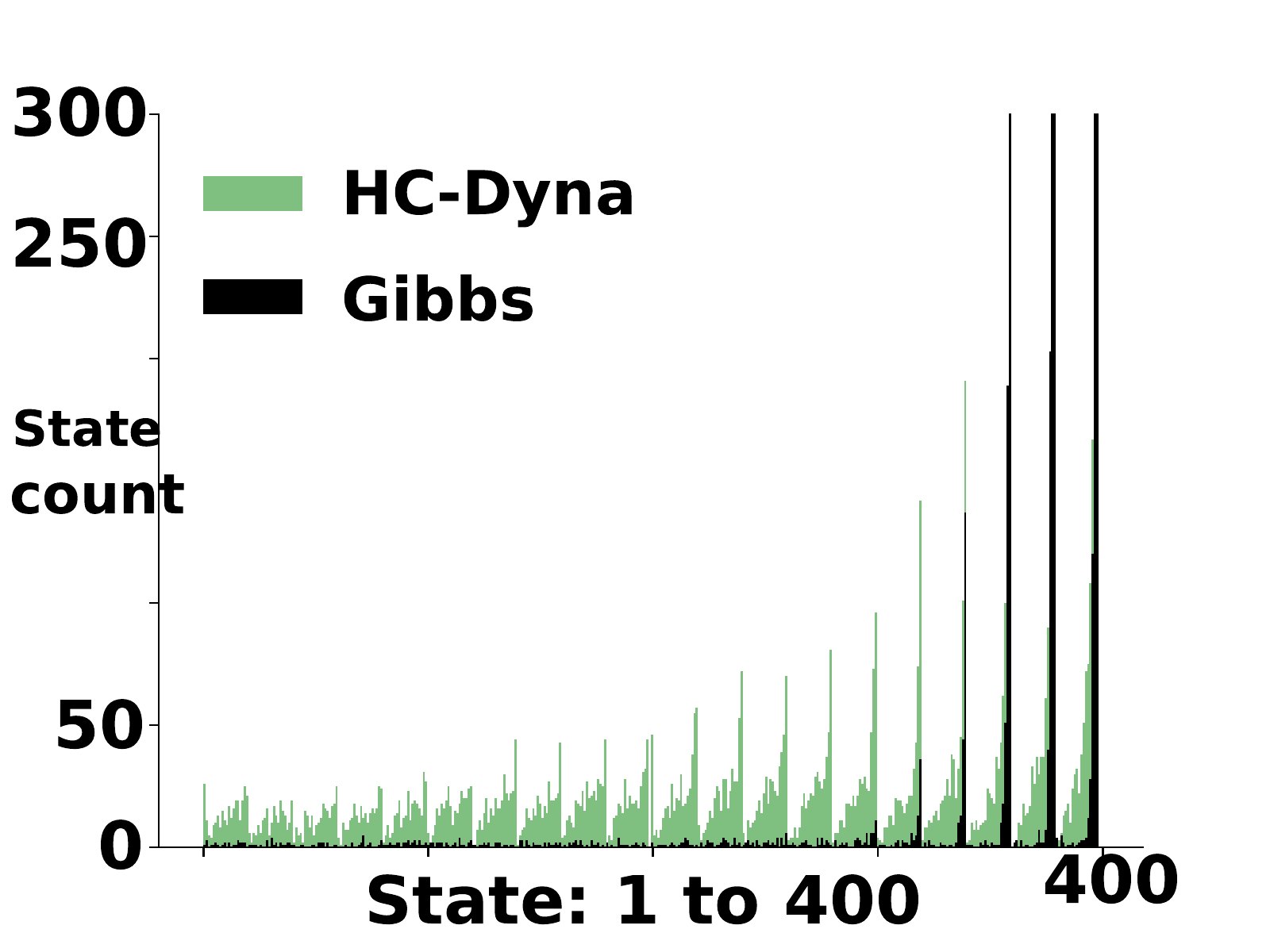} \label{fig:gibbshcdyna_hat_dist}}
		\caption{(a) Evaluation curve; (b) Search-control distribution}\label{fig_sc_trajectory}
	\vspace{-0.25cm}
\end{figure}

As we pointed out the connection to the Langevin dynamics in Section~\ref{sec:HC-Dyna-theory}, the limiting behavior of our hill climbing strategy is approximately a Gibbs distribution. Figure~\ref{fig:tabulargd_compare_scqueue} shows that HC-Dyna performs the best among all sampling distributions, including Gibbs and other baselines. This result suggests that the states during the burn-in period matter.
Figure~\ref{fig:gibbshcdyna_hat_dist} shows the state count by randomly sampling \emph{the same number of states} from the HC-Dyna's search-control queue and from that filled by Gibbs distribution. We can see that the Gibbs one concentrates its distribution only on very high value states. 

\subsubsection{Comparing to True Values}

\begin{figure}[t]
 \vspace{-0.2cm}
	\centering 
		\subfigure[TabularGridWorld]{
			\includegraphics[width=\figwidthfour]{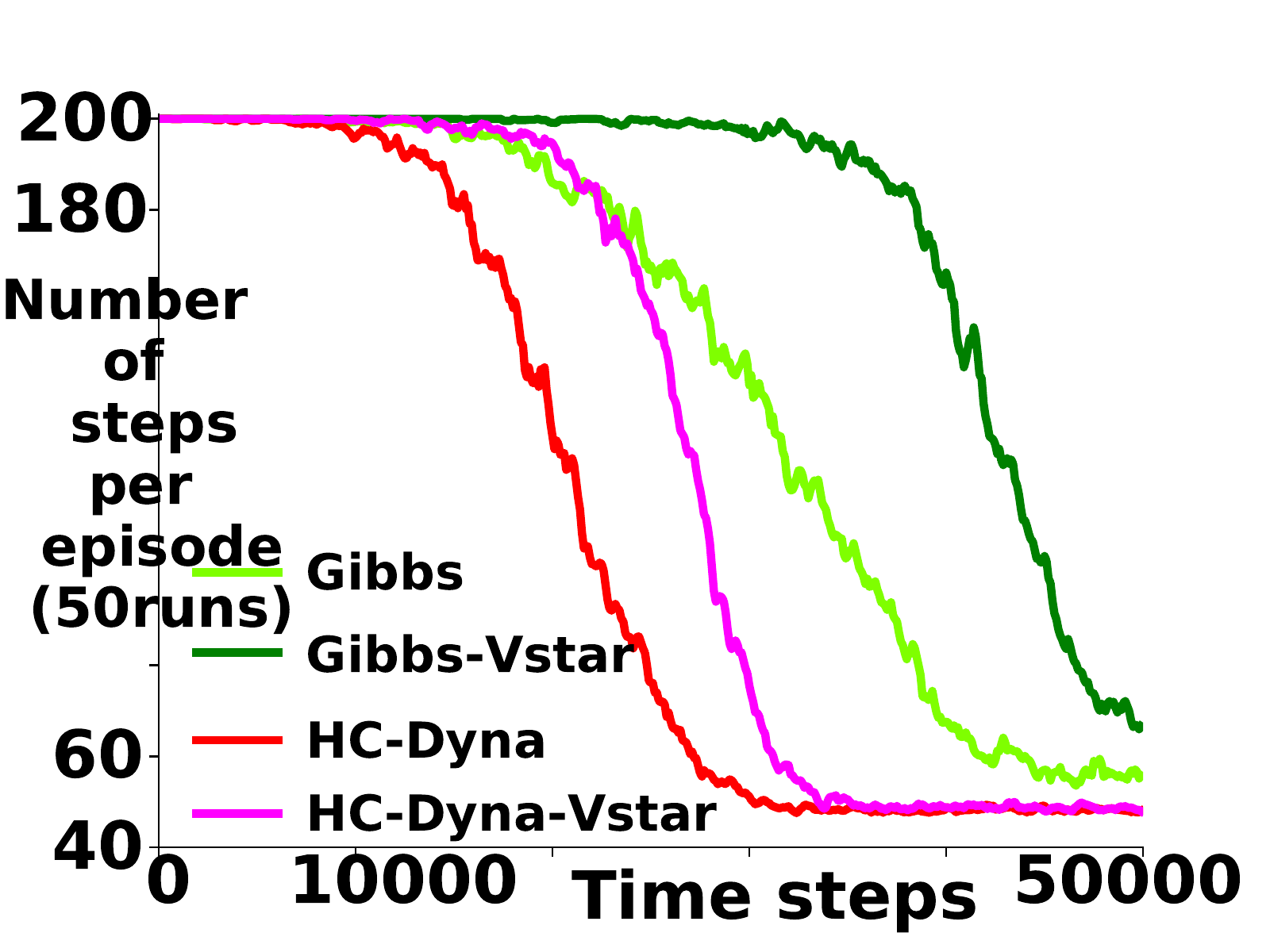}\hfil
			 \label{fig:tabulargd_hat_star}}
		\subfigure[HC-Dyna vs. Gibbs with Vstar]{
			\includegraphics[width=\figwidthfour]{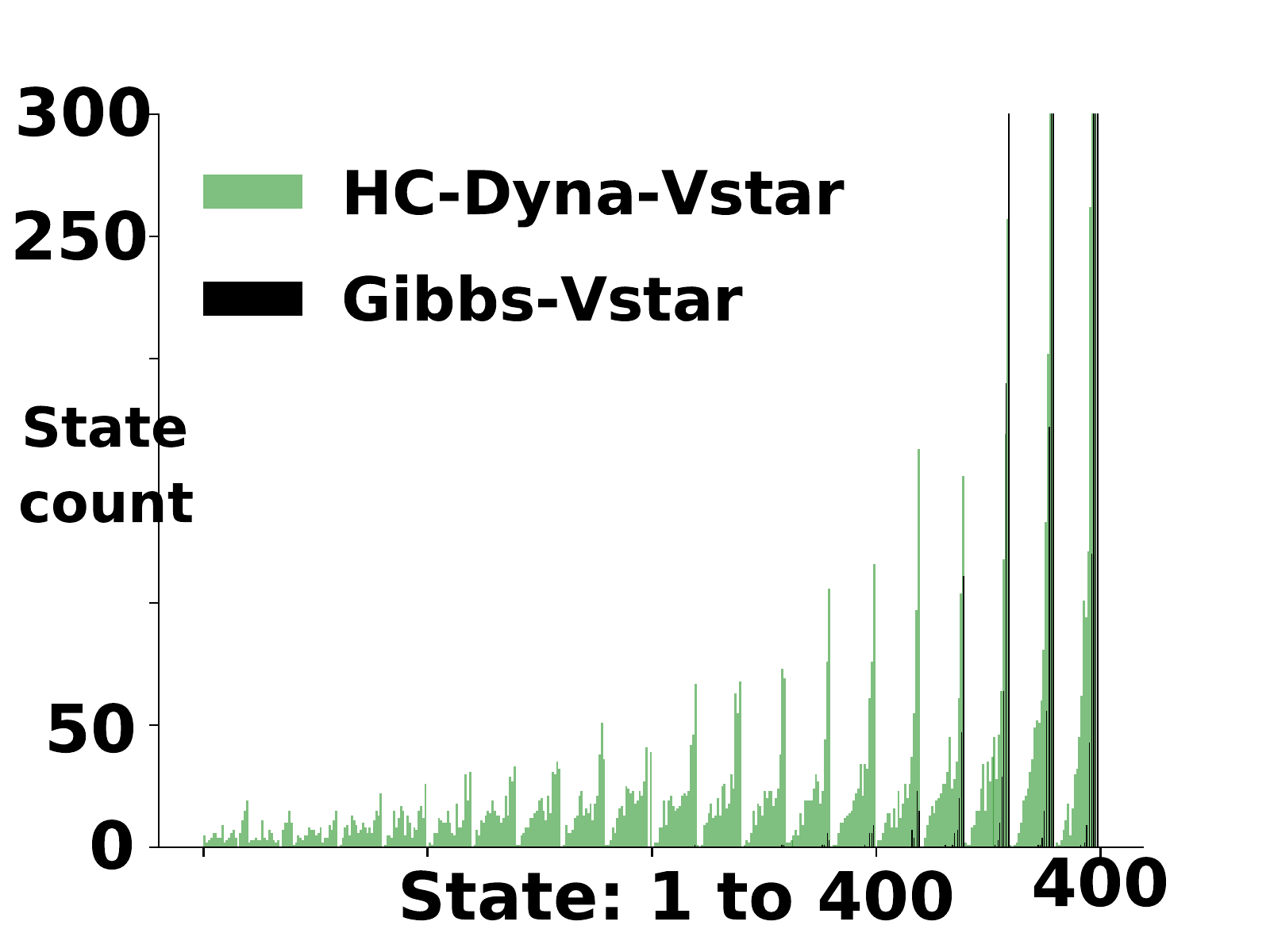} \label{fig:gibbshcdyna_vstar_dist}}
		\caption{(a) Evaluation curve; (b) Search-control distribution}\label{fig_sc_true}
	\vspace{-0.25cm}
\end{figure}

One hypothesis is that the value estimates guide the agent to the goal. 
A natural comparison, then, is to use the optimal values, which should point the agent directly to the goal. 
%
Figure~\ref{fig:tabulargd_hat_star} indicates that using the estimates, rather than true values, is more beneficial for planning. This result highlights that there does seem to be some additional value to focusing updates based on the agent's current value estimates. 
Comparing state distribution of Gibbs-Vstar and HC-Dyna-Vstar in
Figure~\ref{fig:gibbshcdyna_vstar_dist}  to Gibbs and HC-Dyna in Figure~\ref{fig:gibbshcdyna_hat_dist}, one can see that both distributions are even more concentrated, which seems to negatively impact performance. 
 
\section{Conclusion}
\label{sec:HC-Dyna-RL-Conclusion}
We presented a new Dyna algorithm, called HC-Dyna, which generates states for search-control by using hill climbing on value estimates. We proposed a noisy natural projected gradient ascent strategy for the hill climbing process. 
We demonstrate that using states from hill climbing can significantly improve sample efficiency in several benchmark domains. We empirically investigated, and validated, several choices in our algorithm, including the use of natural gradients, the utility of mixing with ER samples, the benefits of using estimated values for search control. A natural next step is to further investigate other criteria for assigning importance to states. Our HC strategy is generic for any smooth function; not only for value estimates.
A possible alternative is to investigate importance based on error in a region, or based more explicitly on optimism or uncertainty, to encourage systematic exploration. 

\section*{Acknowledgments}
We would like to acknowledge funding from the Canada CIFAR AI Chairs Program, Amii and NSERC.


\clearpage
\newpage
\appendix
\section{Appendix}
\label{sec:HC-Dyna-Appendix}

The appendix includes all algorithmic and experimental details. 

\subsection{Algorithmic details}

We include the classic Dyna architecture~\cite{sutton1991dyna,SuttonBarto2018} in Algorithm~\ref{alg_dyna} and our algorithm with additional details in Algorithm~\ref{alg_hcdyna_details}. 

\begin{algorithm}
	\caption{Generic Dyna Architecture: Tabular Setting}
	\label{alg_dyna}
	\begin{algorithmic}
		\STATE Initialize $Q(s,a)$ and model $\mathcal{M}(s, a)$, $\forall (s, a) \in \States \times \Actions$
		\STATE At each learning step: 
		\STATE Receive $s, a, s', r$
		\STATE observe $s$, take action $a$ by $\epsilon$-greedy w.r.t. $Q(s, \cdot)$
		\STATE execute $a$, observe reward $R$ and next state $s'$
		\STATE Q-learning update for $Q(s, a)$
		\STATE update model $\mathcal{M}(s, a)$ (i.e. by counting)
		\STATE put $(s, a)$ into search-control queue
		\FOR{i=1:n}
		\STATE sample $(\tilde{s}, \tilde{a})$ from search-control queue 
		\STATE $(\tilde{s}', \tilde{R}) \gets \mathcal{M}(\tilde{s}, \tilde{a})$ // simulated transition
		\STATE Q-learning update for $Q(\tilde{s}, \tilde{a})$ // planning update
		\ENDFOR
	\end{algorithmic}
\end{algorithm}

\begin{algorithm}
	\caption{(DQN-)HC-Dyna with additional details}
	\label{alg_hcdyna_details}
	\begin{algorithmic}
		\STATE $B_s$: search-control queue, $B$: the experience replay buffer
		\STATE $\mathcal{M}: \States \times \Actions \rightarrow \States \times \RR$, the environment model
		\STATE $k$: number of HC steps, $\alpha$: gradient ascent step size
		\STATE $\rho \in [0,1]$: mixing factor in a mini-batch, i.e. $\rho b$ samples in a mini-batch are simulated from model
		\STATE $\epsilon_a$: threshold for whether one should add a state to $B_s$
		\STATE $n$: number of planning steps
		\STATE $Q, Q'$: current and target Q networks, respectively
		\STATE $b$: the mini-batch size
		\STATE $\tau$: update target network $Q'$ every $\tau$ updates to $Q$
		\STATE $t \gets 0$ is the time step
		\STATE $n_\tau \gets 0$ is the number of parameter updates
		\STATE $d$: number of state variables, i.e. $\States \subset \RR^d$
		\STATE $\covmathat$ is the empricial covariance matrix of states
		\STATE $\mu_{ss} \gets \mathbf{0} \in \RR^{d\times d}, \mu_s \gets \mathbf{0} \in \RR^d$  \ \ \ (assistant variables for computing empirical covariance matrix, sample average will be maintained for $\mu_{ss}, \mu_s$)
		\STATE $\eta$ is the scalar to scale the empirical covariant matrix
		\STATE $\mathcal{N}$ is the multivariate gaussian distribution used as injected noise
		\STATE To reproduce experiment: $\rho = 0.5, b=32, k=100, \eta = 0.1, \tau = 1000, \alpha = 0.1$, DQN learning rate $0.0001$
		\WHILE {true}
		\STATE Observe $s_t$, take action $a_t$ (i.e. $\epsilon$-greedy w.r.t. $Q$)
		\STATE Observe $s_{t+1}, r_{t+1}$, add $(s_t, a_t, s_{t+1}, r_{t+1})$ to $B$
		\STATE $\mu_{ss} \gets \frac{ \mu_{ss}(t-1) + s_ts_t^\top}{t}, \mu_{s} \gets \frac{ \mu_{s}(t-1) + s_t}{t}$
		\STATE $\covmathat \gets \mu_{ss} - \mu_s \mu_s^\top$
		\STATE $\epsilon_a \gets (1-\beta) \epsilon_a + \beta \frac{||s_{t+1}-s_t||_2}{\sqrt{d}}$ for $\beta = 0.001$
		\STATE // Gradient ascent hill climbing
		\STATE sample $s_0$ from $B$, $\tilde{s} \gets \infty$
		\FOR {$i = 0, \ldots, k$ } 
		\STATE $g_{s_i} \gets \nabla_s V(s_i) = \nabla_s \max_a Q(s_i, a)$
		\STATE $s_{i+1}\! \gets \Pi(s_i + \frac{\alpha}{||\hat{\Sigmamat}_s g_{s_i}||} \hat{\Sigmamat}_s g_{s_i} + X_i), X_i \! \sim \mathcal{N}(0, \eta \hat{\Sigmamat}_\svec)$
		\IF {distance$(\tilde{s}, s_{i+1}) \ge \epsilon_a$}
		\STATE Add $s_{i+1}$ into $\bsc$, $\tilde{s} \gets s_{i+1}$
		\ENDIF		
		\ENDFOR
		\STATE // $n$ planning updates: sample $n$ mini-batches
		\FOR {$n$ times}
		\STATE Sample $\rho b$ states from $B_s$, for each state $\hat{s}$, take action $\hat{a}$ according to $\epsilon$-greedy w.r.t $Q$
		\STATE For each state action pairs $(\hat{s}, \hat{a})$, query the model: $\hat{s}', \hat{r} \gets \mathcal{M}(\hat{s}, \hat{a})$, call this Mini-batch-1
		\STATE // then sample from ER buffer
		\STATE Sample Mini-batch-2 with size $b - \rho b$ from $B$, stack Mini-batch-1,Mini-batch-2 into Mini-batch, hence  Mini-batch has size $b$
		\STATE (DQN) parameter update with mini-batch $b_m$
		\STATE $n_\tau \gets n_\tau + 1$
		\IF {$mod(n_\tau, \tau) == 0$}
		\STATE $Q' \gets Q$
		\ENDIF
		\ENDFOR
		\STATE // update the neural network of the model $\mathcal{M}$ if applicable
		\STATE $t \gets t + 1$
		\ENDWHILE
	\end{algorithmic}
\end{algorithm}

\subsection{Experimental details}
\paragraph{Implementation details of common settings.} The GridWorld domain is written by ourselve, all other discrete action domains are from OpenAI Gym \cite{openaigym} with version $0.8.2$. The exact environment names we used are: MountainCar-v0, CartPole-v1, Acrobot-v1. Deep learning implementation is based on tensorflow with version $1.1.0$ \cite{tensorflow2015-whitepaper}. On all domains, we use Adam optimizer, Xavier initializer, set mini-batch size $b = 32$, buffer size $100$k. All activation functions are ReLU except the output layer of the $Q$-value is linear, and the output layer of the actor network is tanh. The output layer parameters were initialized from a uniform distribution $[-0.0003, 0.0003]$, all other parameters are initialized using Xavier initialization \cite{xavier2010deep}. 

As for model learning, we learn a difference model to alleviate the effect of outliers, that is, we learn a neural network model with input $s_t$ and output $s_{t+1} - s_t$. The neural network has two $64$ units hidden ReLU-layers. The model is learned in an online manner and by using samples from ER buffer with a fixed learning rate as $0.0001$ and mini-batch size $128$ across all experiments. 

\paragraph{Termination condition on OpenAI environments.} On OpenAI, each environment has a time limit and the termination flag will be true if either the time limit reached or the actual termination condition satisfied. However, theoretically we should truncate the return if and only if the actual termination condition satisfied. All of our experiments are conducted by setting discount rate $\gamma = 0.0$ if and only if the actual termination condition satisfied. For example, on mountain car, $\emph{done} = true$ \emph{if and only if} the position$\ge 0.5$. 


\paragraph{Experimental details of TabularGridWorld domain.} The purpose of using the tabular domain is to study the learning performances by using different sampling distribution to fill the search-control queue. Our TabularGridWorld is similar to the continuous state domain introduced in \ref{fig:gridworld} except that we do not have a wall and we introduce stochasticity to make it more representative. Four actions are available and can take the agent to the next $\{up, down, left, right\}$ grid respectively. An action can be executed successfully with probability $0.8$ otherwise a random action is taken.
The TabularGridWorld size is $20\times 20$ and each episode start from left-bottom grid and would terminate if reached the right-top grid or $1$k time steps. The return will not be truncated unless the right-top grid is reached. The discount rate is $\gamma = 1.0$. For all algorithms, we fixed the exploration noise as $\epsilon = 0.2$ and sweep over learning rate $\{2^0, 2^{-0.25}, 2^{-0.5}, 2^{-0.75}, 2^{-1}, 2^{-1.5}, 2^{-2.0}, 2^{-2.5}\} = \{1.0, 0.8409, 0.70711, 0.59460, 0.5, 0.35356, 0.25, 0.17678\}$. We fix using exploration noise $\epsilon = 0.2$ and mixing rate $\rho = 0.5$. We use $10$ planning steps for all algorithms. We evaluate each algorithm every $100$ environment time steps. Parameter is optimized by using the last $20\%$ evaluation episodes to ensure convergence. 

For our algorithm HC-Dyna, we do not sweep any additional parameters. We fix doing $80$ gradient ascent steps per environment time step and the injected noise is gaussian $\mathcal{N}(0, 0.05)$. When adding the noise or using finite difference method for computing gradient, we logically regard the domain as $[0, 1]^2$ and hence each grid is a square with length $1/20 = 0.05$. Specifically, when in a grid, we find its corresponding center's $x,y$ coordinates as its location to add noise. As for gradient ascent with finite difference approximation, given a state $s$, we compute the value increasing rate from each of its $8$ neighbors and pick up the one with largest increasing rate as the next state. That is, $s \gets \argmax_{s'} \frac {\hat{V}(s') - \hat{V}(s)}{||s-s'||}$. Both the search-control queue size and ER buffer size are setted as $1e5$. 

The optimal value function used for HC-Dyna-Vstar and Gibbs-Vstar on this domain is acquired by taking the value function at the end of training ER for $1e6$ steps and averaged over $50$ random seeds. 


\paragraph{Experimental details of continuous state domains.}  All continuous state domain, we set discount rate $\gamma = 0.99$. We set the episode length limit as $2000$ for both GridWorld and MountainCar, while keep other domains as the default setting. We use warmup steps $5000$ for all algorithms. 

For all Q networks, we consistently use a neural network with two $32$ units hidden ReLU-layers. We use target network moving frequency $\tau = 1000$ and sweep learning rate $\{0.001, 0.0001, 0.00001\}$ for vanilla DQN with ER with planning step $5$, then we directly use the same best learning rate $(0.0001)$ for all other experiments. For our particular parameters, we fixed the same setting across all domains: mixing rate $0.5$ and $\epsilon_a$ is sample average, number of gradient steps $k=100$ with gradient ascent step size $0.1$ and queue size $1e6$. We incrementally update the empirical covariance matrix. When evaluating each algorithm, we keep a small noise $\epsilon = 0.05$ when taking action and evaluate one episode every $1000$ environment time steps for each run.

\clearpage
\newpage
{
\bibliographystyle{named}
\bibliography{paper}
}


\end{document}